\documentclass[preprint,12pt]{elsarticle}

\usepackage{changepage}
\usepackage{multirow}
\usepackage{booktabs}
\usepackage{amssymb}
\usepackage{float}
\usepackage{tabularx}
\usepackage{adjustbox}
\usepackage{subcaption}
\usepackage{amsxtra}
\usepackage{hyperref}

\journal{Communications in Transportation Research}
\begin{document}
\begin{sloppypar}

\title{STGAtt: A Spatial-Temporal Unified Graph Attention Network for Traffic Flow Forecasting}

% \author[Zliang,JCui,QZeng,NFilipovic.TGeroski,NFilipovic.TGeroski]{Zhuding Liang, Jianxun Cui\corref{corr}, Qingshuang Zeng, Nenad Filipovic, Tijana Geroski}

\author[Zliang]{Zhuding Liang}

\author[JCui]{Jianxun Cui\corref{corr}}
\cortext[corr]{Corresponding author: Jianxun Cui}
\ead{cuijianxun@hit.edu.cn}

\author[QZeng]{Qingshuang Zeng}
\author[FLiu]{Feng Liu}

\author[NFilipovic.TGeroski]{Nenad Filipovic}
\author[NFilipovic.TGeroski]{Tijana Geroski}

\affiliation[Zliang]{
    organization={Faculty of Computing, Harbin Institute of Technology},
    city={Harbin},
    postcode={150001},
    state={HeiLongJiang},
    country={China}
}

\affiliation[JCui]{
    organization={School of Transportation Science and Engineering, Harbin Institute of Technology},
    city={Harbin},
    postcode={150001},
    state={HeiLongJiang},
    country={China}
}

\affiliation[QZeng]{
    organization={Company of China Software Testing Center (MIIT Software and Integrated Circuit Promotion Center)},
    city={Beijing},
    postcode={102206},
    % state={Beijing},
    country={China}
}

\affiliation[FLiu]{
    organization={Transportation Research Institute (IMOB), Hasselt University},
    city={Hasselt},
    postcode={3500},
    country={Belgium}
}

\affiliation[NFilipovic.TGeroski]{
    organization={Faculty of Engineering, University of Kragujevac},
    city={Kragujevac},
    postcode={34000},
    country={Serbia}
}

\maketitle

\begin{abstract}
Accurate and timely traffic flow forecasting is crucial for intelligent transportation systems. This paper presents a novel deep learning model, the Spatial-Temporal Unified Graph Attention Network (STGAtt). By leveraging a unified graph representation and an attention mechanism, STGAtt effectively captures complex spatial-temporal dependencies. Unlike methods relying on separate spatial and temporal dependency modeling modules, STGAtt directly models correlations within a Spatial-Temporal Unified Graph, dynamically weighing connections across both dimensions. To further enhance its capabilities, STGAtt partitions traffic flow observation signal into neighborhood subsets and employs a novel exchanging mechanism, enabling effective capture of both short-range and long-range correlations. Extensive experiments on the PEMS-BAY and SHMetro datasets demonstrate STGAtt's superior performance compared to state-of-the-art baselines across various prediction horizons. Visualization of attention weights confirms STGAtt's ability to adapt to dynamic traffic patterns and capture long-range dependencies, highlighting its potential for real-world traffic flow forecasting applications.
\end{abstract}


\section{Introduction}

Real-time traffic flow forecast has emerged as a critical component in the development of intelligent transportation systems (ITS). By providing accurate and timely forecasting of traffic conditions, ITS can optimize traffic flow, alleviate congestion, and enhance overall transportation efficiency \cite{njoku2023prospects,figueiredo2001towards,zhang2011data,an2011survey}. The increasing availability of real-time traffic data from various sources, including loop detectors, GPS devices, and crowdsourcing platforms, has created unprecedented opportunities for developing data-driven traffic flow prediction models \cite{liu2022intelligent,choudhary2023enhancing,guo2024explainable}. While time series analysis and machine learning methods have shown promise, they often struggle to capture the complex nonlinear spatial-temporal relationships and long-range dependencies present in real-world traffic data. Consequently, there has been a surge of interest in leveraging the power of deep learning techniques, which excel at automatically learning complex patterns from large-scale datasets \cite{lu2020st,yao2018modeling,wu2024multi}.

Previous works, such as \cite{song2020spatial,yu2023sthsgcn,xia2024spatiotemporal}, have categorized the correlations between nodes (sensors) in traffic flow signals into three types. Temporal correlations capture the dependency of a node's state on its own past states. Spatial correlations describe the influence between different nodes at a given time. Building on these, Spatial-temporal correlations refer to the combined dependency of nodes across both time and space. For example, the traffic condition of a node not only affects its own state in the future (a temporal dependency) but also influences the conditions of its neighboring nodes in the future, demonstrating a classic spatial-temporal relationship.Several methods \cite{liu2017short,li2017diffusion,yu2017spatio,wu2019graph,zhao2019t,guo2019attention,zheng2020gman,xu2020spatial,chen2020multi,huang2022learning} have been proposed to capture these influences. These methods typically design separate spatial and temporal modules to extract temporal and spatial influences, respectively. They then attempt to capture spatial-temporal influences by stacking these modules. However, this indirect dependency capturing approach may lead to the omission of crucial information during certain steps.

To address this limitation, we propose to consider temporal, spatial, and spatial-temporal correlations within a unified framework. We refer to the traffic state of a node at a particular time step in the observed traffic flow signal as a state element. We then define cross-dimensional spatial-temporal connectivity based on the connections between these state elements, as detailed in Section \ref{Spatial-Temporal Unified Graph}. This connectivity allows us to process information from both spatial and temporal dimensions in a unified manner. Additionally, Previous attempts to construct a unified graph across spatial-temporal dimensions, such as STSGCN \cite{song2020spatial}, STFGNN \cite{li2021spatial}, have employed Graph Convolutional Networks (GCNs) to process cross-dimensional connections. However, the GCNs used in these works assign equal weights to all connections, failing to differentiate between connections in the spatial and temporal dimensions. While STHSGCN \cite{yu2023sthsgcn} attempts to address this by utilizing three different types of GCNs to model spatial and temporal connections separately, we believe that attention mechanisms offer a more effective and dynamic approach to handling connections with diverse attributes. This is because attention mechanisms compute data-driven attention scores that function as adaptive weights for graph edges, enabling the model to differentiate the varying influence of connections across spatial and temporal dimensions. Although \cite{feng2022adaptive} employs an attention mechanism to distinguish between cross-dimensional connections, it restricts the receptive field of attention to a local scope in order to efficiently capture the aforementioned three types of correlations.  This limitation hinders the model's ability to perceive global patterns effectively.

To overcome these limitations, we first construct a Spatial-Temporal Unified Graph (see \ref{Spatial-Temporal Unified Graph}). Leveraging the physical structure of the graph, we select several state elements as base nodes (see \ref{sec:base_node_selection}). The state elements within a certain Spatial-Temporal Distance (see \ref{Spatial-Temporal Disstance}) from a base node are considered as its first-order neighbors. These neighbors, along with the base node, constitute a neighborhood set. By repeating this process, the entire traffic flow signal,represented on the Spatial-Temporal Unified Graph, is divided into multiple, mostly non-overlapping neighborhood subsets centered around the selected base nodes. Attention is then computed within each neighborhood set. We name our model the \textbf{S}patial-\textbf{T}emporal Unified \textbf{G}raph \textbf{Att}ention Network (STGAtt).  By computing attention based on the Spatial-Temporal Unified Graph, STGAtt directly captures the underlying correlations across both spatial and temporal dimensions.

As the scale of the traffic network increases, the number of neighborhood subsets partitioning the traffic flow signal inevitably grows. To facilitate information exchange between these sets, we adapt the approach from \cite{liu2021swin}. Specifically, we allow neighboring sets to exchange state elements and perform attention calculations to enable information interaction. Before the next attention computation, we shift the base node of each neighborhood set by a specific distance in both the spatial and temporal dimensions. New neighborhood subsets are then formed around these shifted base nodes using the same partitioning method. This process ensures that adjacent sets exchange approximately half of their state elements before the subsequent attention calculation (see \ref{sec:partition}). Combining the attention results from two consecutive partitionings effectively achieves information exchange across different neighborhood subsets. Through this partitioning and exchange scheme, our STGAtt model effectively captures both short-range and long-range spatial-temporal correlations. In summary, the main contributions of this work are:

\begin{enumerate}
    \item We propose the Spatial-Temporal Unified Graph to provide a unified perspective for traffic flow forecasting. By leveraging an attention mechanism, we effectively utilize the inherent physical connections in the spatial dimension while accommodating the differences in connectivity along the temporal dimension. Treating connections from both dimensions as an integrated whole allows us to effectively capture the underlying spatial-temporal correlations within the traffic data.

    \item Our STGAtt model partitions the traffic signal into multiple independent state element neighborhood subsets based on the Spatial-Temporal Unified Graph. Attention is computed independently within each subset. Furthermore, by continuously exchanging state elements between subsets and performing attention calculations, STGAtt enables inter-subset information interaction, capturing both short-range and long-range spatial-temporal correlations.

    \item Extensive experiments conducted on real-world traffic datasets demonstrate that our model achieves competitive prediction performance compared to existing baselines.

\end{enumerate}

\section{Related Works}
Predicting traffic flow is a fundamental challenge in intelligent transportation systems, crucial for applications like traffic control, guidance, emergency management, and decision-making \cite{zhu2022kst,sayed2023artificial,xing2023overview,navarro2022traffic}. Broadly categorized, traffic flow prediction methods can be classified into model-driven and data-driven paradigms \cite{zhao2019t,zhang2023urban,mrad2023overview}. While model-driven approaches primarily rely on time series analysis, data-driven methods encompass traditional machine learning and deep learning techniques \cite{guo2019attention,huang2022modern}.

\subsection{Model-Driven Approaches}
Early traffic flow prediction research heavily utilized time series analysis methods like ARIMA \cite{hamed1995short}, KARIMA \cite{van1996combining}, SARIMA \cite{williams2003modeling}, and Kalman filtering \cite{okutani1984dynamic}. While simple, these methods often assume stationarity, failing to capture the inherent nonlinearity and uncertainty of real-world traffic patterns, leading to less accurate predictions.

\subsection{Data-Driven Methods}
To address this, researchers explored non-parametric traditional machine learning methods such as K-Nearest Neighbors \cite{zhang2013improved}, Support Vector Regression (SVR) \cite{nidhi2022traffic}, Fuzzy Logic \cite{chen2018novel}, Bayesian Networks \cite{pascale2011adaptive}, and Neural Networks \cite{wang2021traffic}. While these methods offered improved capacity to model complex traffic data relationships, their ability to effectively leverage the vast amounts of available traffic data remained limited.

Deep learning techniques have been used to capture spatial-temporal dependencies to predict traffic flow. These methods normally employ separate spatial-temporal modeling to handle temporal and spatial dependencies.Recurrent Neural Networks (RNNs) like Long Short-Term Memory (LSTM) \cite{liu2017short} and gated recurrent unit (GRU) \cite{li2017diffusion}, as well as 1D-CNN and Temporal Convolutional Network (TCN) \cite{wu2019graph}, are commonly used for capturing temporal features. For spatial dependencies, Graph Convolutional Networks (GCNs) are prevalent choice \cite{yu2017spatio,wu2019graph,zhao2019t}. However, this separation struggles to capture the intricate interplay between spatial and temporal aspects of traffic flow. Consequently, researchers integrated attention mechanisms alongside or in place of RNNs, CNNs, and GCNs within spatial-temporal modules to enhance the capture of spatial-temporal correlations \cite{guo2019attention,zheng2020gman,xu2020spatial,chen2020multi,huang2022learning}.

\subsection{Unified Graph Representations}
Recognizing the limitations of separate modeling, recent research focuses on building unified graph representations to capture the interconnectedness of spatial and temporal information using Graph Convolutional Networks (GCNs). For instance, STSGCN \cite{song2020spatial} constructs a localized spatial-temporal graph for learning through a GCN. Building on this, STFGNN \cite{li2021spatial} employs Dynamic Time Warping (DTW) \cite{muller2007dynamic} to align node embeddings and fuses multiple spatiotemporal graphs while integrating temporal convolutions. ASTTN \cite{feng2022adaptive} directly models cross-time correlations using a localized multi-head self-attention mechanism on the spatial-temporal graph. STHSGCN \cite{yu2023sthsgcn} leverages three distinct graph convolutions working synchronously to capture spatial-temporal heterogeneities.

Despite these advancements, challenges remain. GCN-based methods often struggle to dynamically differentiate between spatial and temporal connections, while localized attention mechanisms may overlook crucial global dependencies. To overcome these limitations, we introduce STGAtt, a novel Spatial-Temporal Unified Graph Attention Network designed to effectively capture both short-range and long-range spatial-temporal dependency in traffic flow forecasting. The essential differences between STGAtt and the existing methods, particularly "unified methods", lie in the following aspects:

\begin{itemize}
    \item \textbf{Dynamic Weighting Mechanism:} While existing unified methods often rely on GCNs which assign static or pre-defined weights to connections (e.g., STSGCN \cite{song2020spatial}) or use a limited set of GCN types (e.g., STHSGCN \cite{yu2023sthsgcn}), STGAtt employs a self-attention mechanism. This allows the model to compute data-driven, dynamic weights for connections on the unified graph, enabling it to more effectively differentiate the varying importance of influences from different spatial and temporal neighbors.

    \item \textbf{Global Information Exchange with Local Computation:} Unlike localized attention mechanisms that restrict the receptive field and hinder the capture of global patterns (e.g., ASTTN \cite{feng2022adaptive}), STGAtt introduces a novel partitioning and shifting strategy. By first computing attention locally within partitioned neighborhood subsets for efficiency, and then systematically shifting these sets to enable information exchange between them, our model effectively captures both short-range and long-range spatial-temporal correlations, achieving a global receptive field without incurring prohibitive computational costs.
\end{itemize}

\section{Methodology}
\subsection{Preliminaries}
\subsubsection{Traffic Flow Forecast}
\label{Traffic Flow Forecast}
Given a traffic network represented as a graph $G = (\nu ,\varepsilon ,A)$, where $\nu$ is the set of nodes with $\left| \nu  \right| = N$ representing the number of nodes, $\varepsilon$ is the set of edges with $\left| \varepsilon  \right| = E$ representing the number of edges, and $A \in {\mathbb{R}^{N \times N}}$ is the adjacency matrix describing the connectivity between the $N$ nodes. At a specific time $t$, the observed traffic flow across all nodes over the past $T$ time steps can be represented as a signal $X \in {\mathbb{R}^{N \times T \times C}}$, where $C$ is the dimension of the traffic state, such as flux, density, or occupancy. The task of traffic flow forecasting is to learn a mapping function $f$ based on $G$ and $X$ that predicts the traffic flow signal $\hat X \in {\mathbb{R}^{N \times T' \times C}}$ for the future $T'$ time steps. This relationship can be expressed as:

\begin{equation}
    [X;G]\xrightarrow{f}\hat X
\end{equation}

\subsubsection{Spatial-Temporal Unified Graph}
\label{Spatial-Temporal Unified Graph}

\begin{figure}[htbp]
    \centering
    \includegraphics[width=0.6\textwidth]{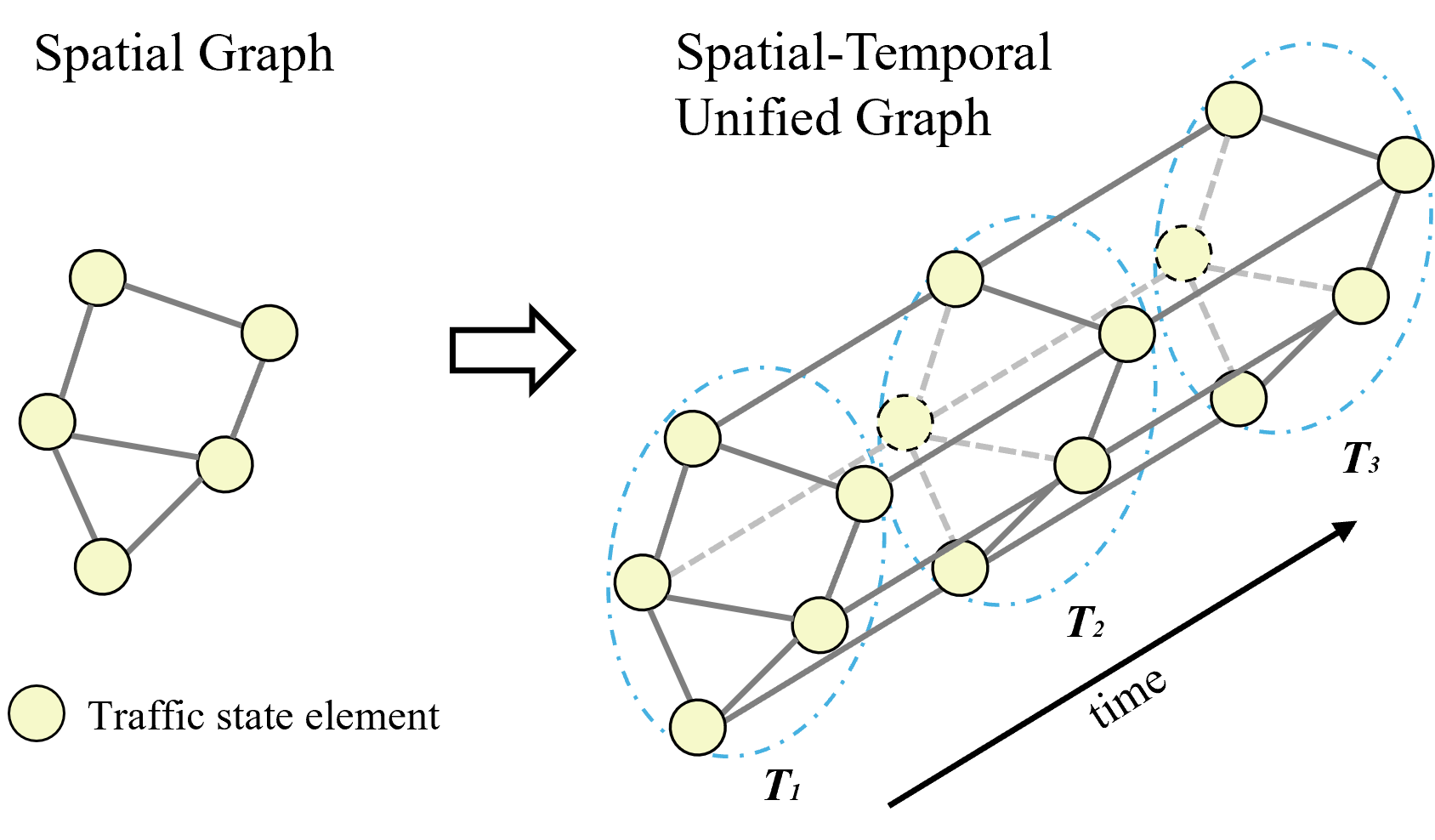}
    \caption{Illustration of Spatial-Temporal Unified Graph construction. First, spatial relationships are captured in a spatial graph (left). Then, this structure is replicated across multiple time steps ($T_1$, $T_2$, $T_3$) and augmented with temporal edges connecting identical nodes between adjacent time steps, resulting in the unified graph (right).}
    \label{fig:Spatial-Temporal Unified Graph}
\end{figure}

For a traffic flow signal $X \in {\mathbb{R}^{N \times T \times C}}$, we aim to represent its spatial and temporal connections from a unified perspective. We define ${X_{i,j}} \in {\mathbb{R}^C}$ as the traffic state element representing the traffic state of the $i$-th node at the $j$-th time step in $X$. In the spatial dimension, the connectivity between state elements ${X_{k,t}}$ and ${X_{k',t}}$ within the same time step can be described by the connectivity between nodes ${N_k}$ and ${N_{k'}}$, which is defined in the adjacency matrix $A$. Extending this connectivity to the temporal dimension, we introduce the Spatial-Temporal Unified Adjacent Matrix ${A^{ST}} \in {\mathbb{R}^{NT \times NT}}$ based on the given adjacency matrix $A$:

\begin{equation}
    \label{sts_adj_matrix}
    A_{_{((i,j),(i',j'))}}^{ST} = 
    \begin{cases}
        1, &  i = i', \left| j - j' \right| = 1 \\
	A_{i,i'}, &  j=j' \\
	0, & otherwise
    \end{cases}
\end{equation}
Here, $A_{_{((i,j),(i',j'))}}^{ST}$ represents the connectivity between state elements ${X_{i,j}}$ and ${X_{i',j'}}$. $A_{i,i'}$ is the value of the $i$-th row and $i'$-th column in the adjacency matrix $A$. This piecewise function is designed to unify spatial and temporal adjacencies into a single matrix. The logic is as follows:
\begin{itemize}
    \item The first case ($i = i', \left| j - j' \right| = 1$): This rule establishes connections in the temporal dimension. It states that for the same spatial node ($i = i'$), there is a direct link between its states at two adjacent time steps ($|j - j'| = 1$). This captures the continuity of traffic flow, where the state of a node at time $t$ directly influences its state at time $t+1$. In Fig. \ref{fig:Spatial-Temporal Unified Graph}, these correspond to the edges connecting the same node across different time steps.

    \item The second case ($j=j'$): This rule establishes connections in the spatial dimension. It states that within the same time step ($j = j'$), the connectivity between different nodes ($i$ and $i'$) is determined by the original spatial adjacency matrix $A$. This effectively replicates the spatial graph structure at each time step, capturing the influence between physically connected road segments at a specific moment. In Fig. \ref{fig:Spatial-Temporal Unified Graph}, these are the edges within each time step (e.g., within $T_1$ or $T_2$).
\end{itemize}

By integrating spatial and temporal connectivity, we construct a unified perspective of a spatial-temporal unified graph ${G^{ST}} = ({\nu ^{ST}},{\varepsilon ^{ST}},{A^{ST}})$, where $\nu ^{ST}$ is the set of nodes with  $\left| {\nu ^{ST}}  \right| = NT$ from the unified spatial-temporal perspective, $\varepsilon ^{ST}$ is the set of edges with $\left| {\varepsilon ^{ST}}  \right| < (NT)^2$, and ${A^{ST}}$ is the Spatial-Temporal Unified Adjacent Matrix defined by Eq.\ref{sts_adj_matrix}. Fig.\ref{fig:Spatial-Temporal Unified Graph} shows an example of Spatial-Temporal Unified Graph.

\subsubsection{Spatial-Temporal Distance}
\label{Spatial-Temporal Distance}
Given two state elements ${X_{i,j}},{X_{i',j'}} \in {\mathbb{R}^{C}}$ from the traffic flow signal $X$, we define their Spatial-Temporal Distance in a simplified manner as:

\begin{equation}
    \label{eq:st_diss}
    D^{ST} \left( {{X_{i,j}},{X_{i',j'}}} \right) = \min \{ p \in \mathbb{N}|{ \left( (A^{ST})^p \right)_{((i,j),(i',j'))}} > 0\}
\end{equation}
where $( \cdot )^p $ denotes the $p$-th power of a matrix. $ ( A^{ST} )_{((i,j),(i',j'))} $ denotes the element at the row corresponding to state element ${X_{i,j}}$ and the column corresponding to state element ${X_{i',j'}}$ in the adjacency matrix ($A^{ST}$). This definition leverages a fundamental principle of graph theory to precisely define the spatial-temporal distance as the shortest path length between two nodes on the Spatial-Temporal Unified Graph $G^{ST}$. In graph theory, a non-zero entry in the $p$-th power of the adjacency matrix, $(A^{ST})^p$, signifies the existence of a path of length $p$ between two nodes. Therefore, finding the minimum $p$ that satisfies this condition is equivalent to calculating the geodesic distance between them.

% This definition simplifies the calculation of spatial-temporal distance between ${X_{i,j}}$ and ${X_{i',j'}}$ by utilizing an estimated distance derived from the Spatial-Temporal Unified Adjacent Matrix $A^{ST}$.

\subsection{Framework of STGAtt}

\begin{figure}[H]
    \centering
    \includegraphics[width=1\textwidth]{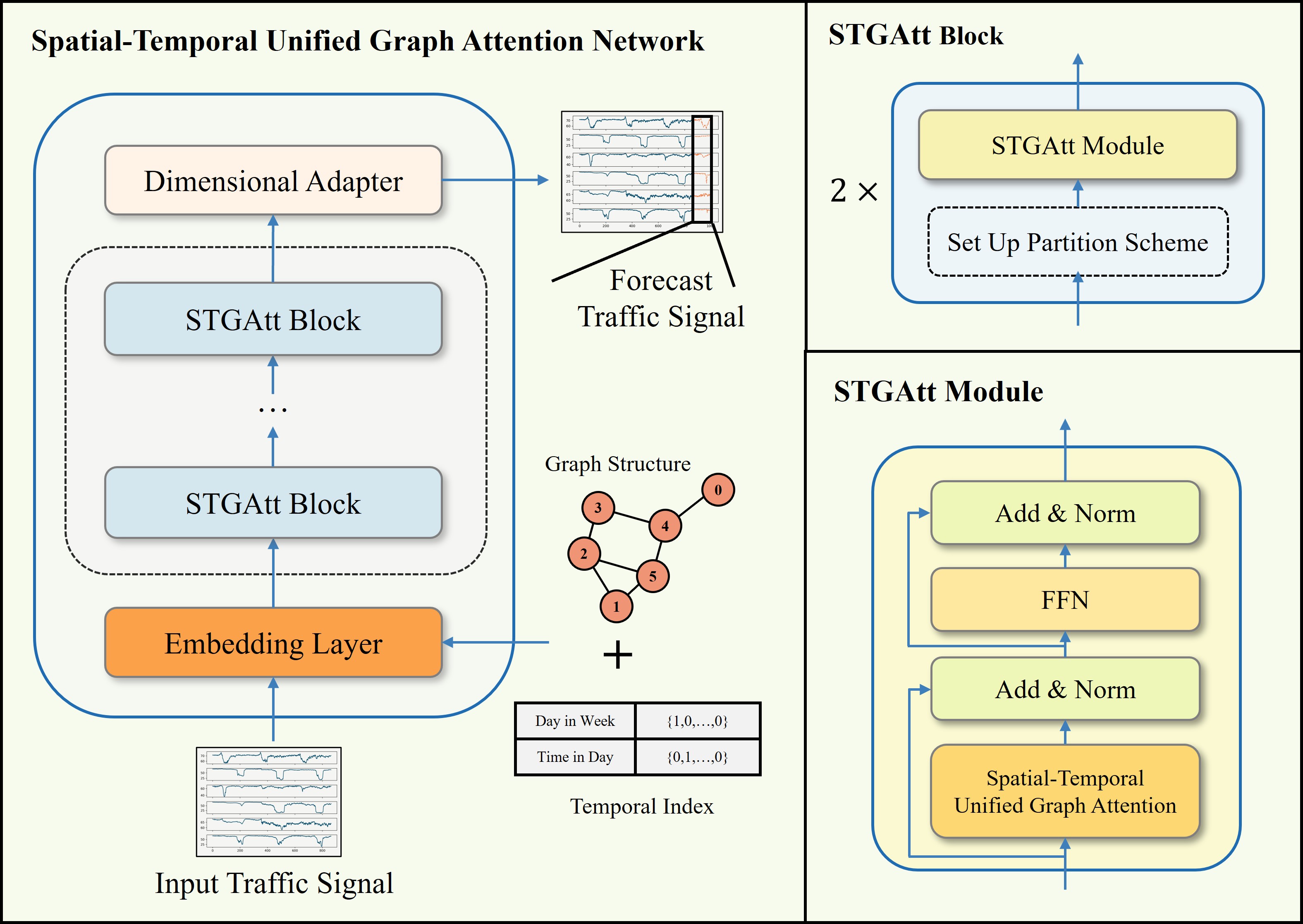}
    \caption{The framework of STGAtt. The STGAtt takes traffic signal, graph structure, and temporal index as input. It consists of an embedding layer, multiple STGAtt blocks, and a dimensional adapter. The embedding layer maps raw data to a latent space and adds spatial and temporal positional encodings. Each STGAtt block captures spatial-temporal correlations in traffic flow data using a Spatial-Temporal Unified Graph Attention mechanism. It partitions data into subsets of state elements, computes attention within subsets, and exchanges elements between subsets for information interaction through two pre-defined partition schemes (i.e. P1 and P2 which are illustrated in \ref{sec:partition}). The dimensional adapter transforms the final latent representation to the forecast traffic signal output.}
    \label{fig:STGAtt_Architecture}
\end{figure}

To predict future traffic flow $\hat X \in {\mathbb{R}^{N \times T' \times C}}$ given a traffic network $G$ and historical traffic flow data $X \in {\mathbb{R}^{N \times T \times C}}$, the STGAtt processes the data through a series of steps as illustrated in Fig.\ref{fig:STGAtt_Architecture}. 

First, the raw traffic flow data $X$ is fed into an \textbf{Embedding Layer}. This layer first maps the input data to a latent space using a linear layer. Then, spatial position embedding (SPE) and temporal position embedding (TPE) are added to the output of the first step of this layer. The SPE is calculated based on the Laplacian eigenvectors of the network's adjacency matrix $A$, while the TPE is generated based on the day-in-week and time-in-day of each traffic state element. Finally, another linear layer and a normalization layer are applied to output the traffic flow embedding $\tilde X \in {\mathbb{R}^{N \times T \times D}}$.

The core of the STGAtt is the \textbf{STGAtt Block}, which is responsible for capturing the spatial and temporal correlations within the traffic data. This block utilizes a novel approach to dynamically weigh connections between traffic state elements in a \textbf{Spatial-Temporal Unified Graph}. For $\forall h \in [1,H]$ STGAtt Block, its input and output can be represented as ${{\tilde X}^{(h - 1)}} \in {\mathbb{R}^{N \times T \times D}}$ and ${{\tilde X}^{(h)}} \in {\mathbb{R}^{N \times T \times D}}$ separately. Specially, ${{\tilde X}^{(0)}} = {Embedding Layer}(X)$.

Each STGatt Block has two identical and serially connected STGatt modules, but before the data enters each of them, one of a pair of data partition scheme (i.e. P1 Partition and P2 Partition which is described in detail in \ref{sec:partition}) should be implemented first. This scheme, derived from the \textbf{Spatial-Temporal Distance} using the Spatial-Temporal Unified Adjacent Matrix ${A^{ST}}$, divides the data into several neighborhood subsets. In each neighborhood set partitioned by P1 and P2, state elements are spatially and temporally close. By shifting the base nodes in P1 scheme, P2 scheme will lead to a different neighborhood set partition layout, but with overlap of each  pair of corresponding partitioned sets of P1 and P2. These overlaps of state elements can act as "brige nodes" to faciliate information exchange between neighborhood subsets when we conduct attention mechanism calculations on each standalone neighborhood set.

Within each STGAtt Module, the \textbf{Spatial-Temporal Unified Graph Attention} mechanism is applied to each neighborhood set individually. This mechanism calculates the attention weights between each state elements within the neighborhood set, allowing the network to learn complex dependencies. Following the attention mechanism, a Feed-Forward Networks (FFN) layer is applied, and skip connections and normalization are employed to enhance the learning process.

Finally, the output of the last STGAtt Block is processed by the \textbf{Dimensional Adapter}. This module consists of two linear transformations. The first transformation adjusts the temporal dimension of the latent traffic flow representation to match the desired prediction horizon $T'$. The second transformation maps the adjusted latent representation to the final prediction output $\hat X$ in the original data space. Detailed module illustrations are given below.

\subsubsection{Embedding Layer}
Since we will use an attention-based model to capture the spatial and temporal correlations in traffic flow data, it is important to add position encoding to ensure the uniqueness of each input token in time and space, also to preserve the relative distance information between each pair of input traffic state elements \cite{ying2021transformers,zhang2024dynamic}. As illustrated in Fig.\ref{fig:STGAtt_Architecture}, we first project the raw traffic flow signal $X \in \mathbb{R}^{N \times T \times C}$ into a latent space using a linear layer. Subsequently, we incorporate spatial position embedding (SPE) and temporal position embedding (TPE). Finally, the output is processed by another linear layer and a normalization layer, resulting in the traffic flow embedding $\tilde{X} \in \mathbb{R}^{N \times T \times D}$, which serves as the output of the Embedding Layer. This process can be summarized as follows:

\begin{equation}
    \begin{gathered}
      \tilde X = Linear(X) \hfill \\
      \tilde X = \tilde X{ \oplus _{broadcast}}SPE{ \oplus _{broadcast}}TPE \hfill \\
      \tilde X = Linear(\tilde X) \hfill \\
      \tilde X = Norm(\tilde X) \hfill \\ 
    \end{gathered} 
\end{equation}
where $\oplus _{broadcast}$  is the tensor plus operation based on broadcasting mechanism. 

\textbf{Spatial Position Embedding (SPE): }In the spatial dimension, we adopt the position encoding method proposed in \cite{dwivedi2020generalization} to embed the graph structure. We begin by computing the Laplacian eigenvectors of the adjacency matrix $A$ corresponding to the traffic graph $G$:

\begin{equation}
    L = I - {D^{ - 1/2}} A {D^{1/2}} = {U^T}\Lambda U
\end{equation}
where $L$ denotes the Laplacian matrix, $D$ represents the degree matrix, and $\Lambda$ and $U$ corresponds to the eigenvalues and eigenvectors, respectively. Subsequently, we select the $R (R < N)$ smallest non-trivial eigenvectors for each node as its position encoding. These selected eigenvectors are then fed into a linear layer to obtain the spatial position embedding $SPE \in {\mathbb{R}^{N \times {D}}}$ of the road graph $G$.

\textbf{Temporal Position Embedding (TPE): }In the temporal dimension, referring to the method proposed in \cite{zheng2020gman}, temporal position encoding is generated for each traffic state elements token according to the day-in-week and time-in-day. The "day-in-week" position of each elements is represented by a one-hot encoding vector ${\mathbb{R}^{7}}$ and its "time-in-day" is represented by a one-hot encoding vector ${\mathbb{R}^{\gamma}}$ (where $\gamma$ is the total time steps in one day). Then "day-in-week" position vector and "time-in-day" vector are concatenated to feed into a linear layer to output the temporal position embedding of this traffic state element with shape $TPE \in {\mathbb{R}^{T \times {D}}}$.

\subsubsection{STGAtt Block}
In this block, we utilize STGAtt Module with multi-head attention mechanism to capture spatial, temporal and spatial-temporal dependency in an uniform manner based on the Spatial-Temporal Unified Graph representation. We consider the physical structure of the traffic signal by partitioning it into subsets of state elements based on their Spatial-Temporal Distance. This allows us to compute intra-group attention within each subset, accounting for the physical relationships between state elements. Furthermore, inspired by the Shifted Window mechanism \cite{liu2021swin}, we dynamically adjust the partitioning of state element neighborhood subsets between consecutive attention calculations. This ensures that neighboring subsets exchange approximately half of their state elements, facilitating information exchange across subsets through the stacked attention modules.

\begin{figure}[t]
    \centering
    \includegraphics[width=1\textwidth]{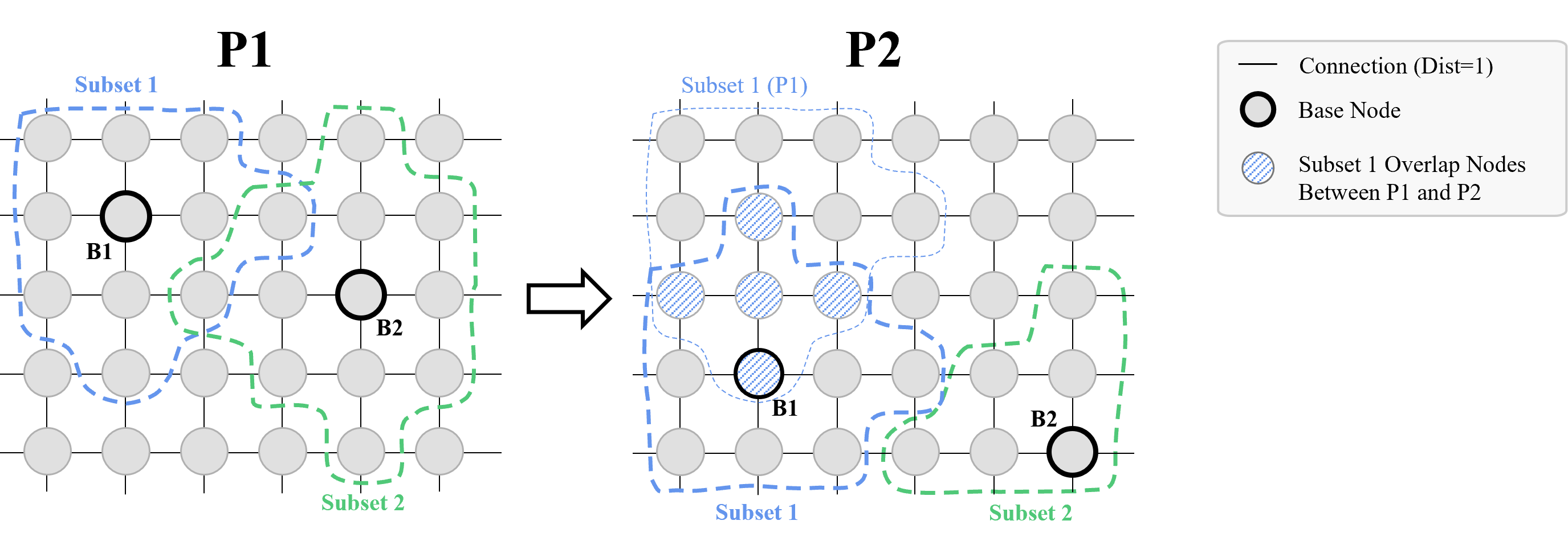}
    \caption {Example of constructing state element neighborhood subset partitions P1 and P2 on a Spatial-Temporal Unified Graph (simplified to a grid), using a distance threshold $\tau=2$. P1 (left) shows state element neighborhood subsets (Subset 1, Subset 2) defined around base nodes B1 and B2 according to Eq. \ref{eq:partition_diss_req}. P2 (right) is derived by shifting the base nodes of P1. The resulting partitions in P2 intentionally overlap with P1's partitions (shown for Subset 1 with hatched nodes representing intersection) to enable information exchange across subsets as described in the text.}
    \label{fig:shift_the_base_nodes}
\end{figure}  

\label{sec:partition}
\textbf{Partitioning of State Element Neighborhood Subsets: } Given the embedding $\tilde X \in {\mathbb{R}^{N \times T \times D}}$ of a traffic flow signal for a specific traffic graph $G$ with $N$ physical nodes and $T$ time steps, we first select a set of $l$ state elements $\{ {{\tilde X}_{{i_1},{j_1}}},{{\tilde X}_{{i_2},{j_2}}}, \ldots ,{{\tilde X}_{{i_l},{j_l}}} \}$ as base nodes, based on the physical structure of the graph (see \ref{sec:base_node_selection}). For each base node, we define a state element neighborhood subset:

\begin{equation}
    \label{eq:partition_diss_req}
    {{\breve X}_{i,j}} = \left \{ {X_{x,y}}|D^{ST}\left( {{\tilde X_{i,j}},{\tilde X_{x,y}}} \right) \leqslant \tau \right \}
\end{equation}
where $\tilde X_{i,j}$ is the given base node and $\tau$ is a predefined threshold. All state elements within ${{\breve X}_{i,j}}$ are considered first-order neighbors of the base node ${\tilde X_{i,j}}$, as shown in Fig.\ref{fig:shift_the_base_nodes}. We strive to ensure that all the $l$ state element neighborhood subsets cover all state elements while maintaining similar subset sizes and minimizing overlap. This means that the set $P1 = \{ \breve X_{{i_1},{j_1}},\breve X_{{i_2},{j_2}}, \ldots ,\breve X_{{i_l},{j_l}} \}$, 
representing the state element neighborhood subsets, should satisfy the following constraints:

\begin{equation}
    \label{eq:partition_set_req}
    \begin{gathered}
        \begin{cases}
            \bigcup\nolimits_{p = 1}^l {{{\breve X}_{{i_p},{j_p}}}}  = \breve S \hfill \\
            \left| {{{\breve X}_{{i_1},{j_1}}}} \right| \approx \left| {{{\breve X}_{{i_2},{j_2}}}} \right| \approx  \ldots  \approx \left| {{{\breve X}_{{i_l},{j_l}}}} \right| \hfill \\
            minimize \sum\nolimits_{p = 1}^l {\sum\nolimits_{q = 1}^l {\left| {{{\breve X}_{{i_p},{j_p}}} \cap {{\breve X}_{{i_q},{j_q}}}} \right|} }  \hfill \\
        \end{cases}
    \end{gathered}
\end{equation}
where $\breve S$ represents the set of all state elements. This defines the $l$ state element neighborhood subsets.

To facilitate information exchange between different state element neighborhood subsets, we enable the subsets to dynamically exchange elements. This allows information to flow across subsets through the stacked attention modules. As illustrated in Fig.\ref{fig:shift_the_base_nodes}, we shift the base nodes in $P1$ by a specific Spatial-Temporal Distance on the Spatial-Temporal Unified Graph. Using these shifted base nodes, we partition the state elements into new subsets following the constraint defined in Eq.\ref{eq:partition_diss_req}. This new partitioning scheme is denoted as $P2 = \{ \breve X_{{i_1}',{j_1}'},\breve X_{{i_2}',{j_2}'}, \ldots ,\breve X_{{i_l}',{j_l}'} \}$. Similar to $P1$, the subsets in $P2$ are required to satisfy the constraints in Eq.\ref{eq:partition_set_req}. Additionally, we enforce the following constraint on $P2$:

\begin{equation}
    \begin{gathered}
        \forall p \in \{ 1,2, \ldots ,l \},  \hfill \\
        \begin{cases}
            \left| \breve X_{{i_p}',{j_p}'} \cap \breve X_{{i_p},{j_p}} \right| \approx \frac{1}{2}\left| \breve X_{{i_p},{j_p}} \right| \hfill \\
            \left| \breve X_{{i_p}',{j_p}'} \cap \left( \breve S \setminus \breve X_{{i_p},{j_p}} \right) \right| \approx \frac{1}{2}\left| \breve X_{{i_p},{j_p}} \right| \hfill \\
        \end{cases}
    \end{gathered}
\end{equation}
This constraint ensures that each new subset $\breve X_{i',j'} \in P2$ contains approximately half of its elements from its corresponding subset $\breve X_{i,j} \in P1$ and the other half from other subsets, effectively promoting information exchange.

\begin{figure}
    \centering
    \includegraphics[width=0.5\textwidth]{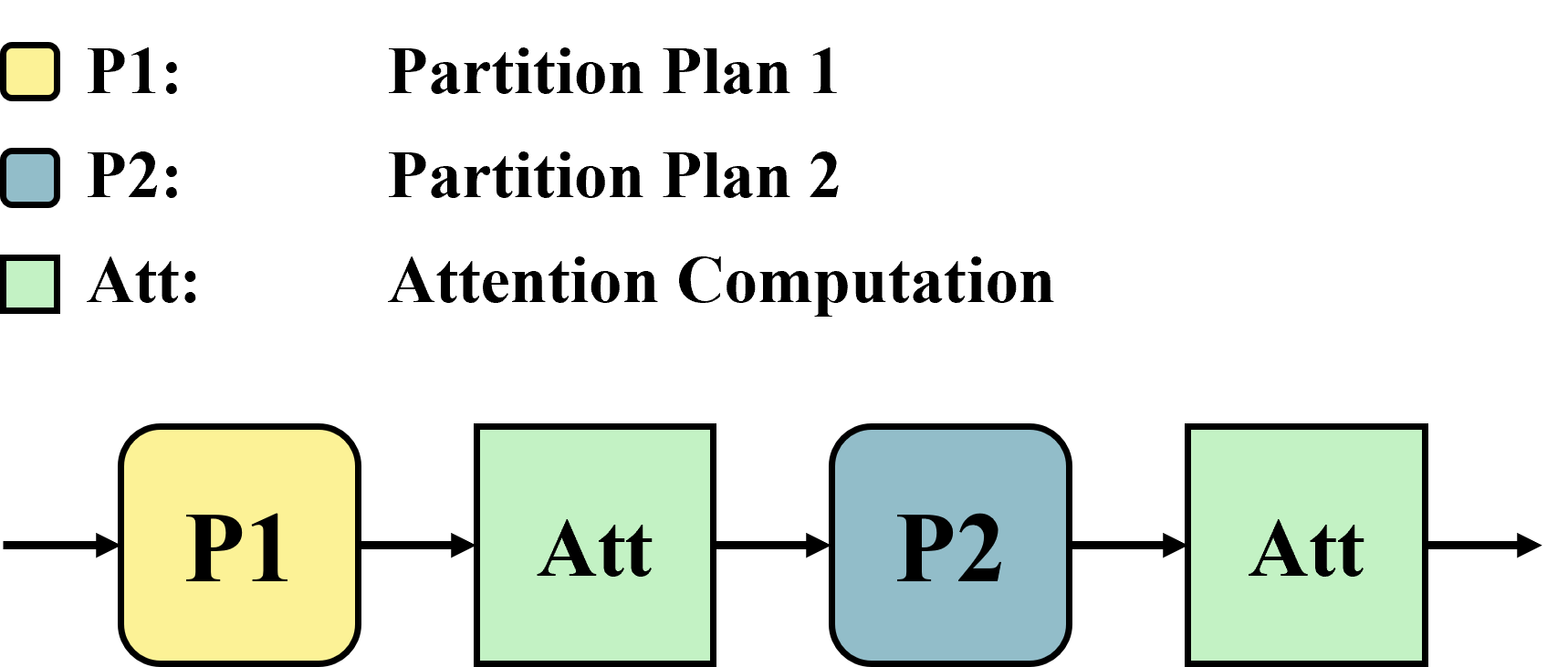}
    \caption {
         Cyclic utilization of the partitioning schemes $P1$ and $P2$.
    }
    \label{fig:two_stage_partition}
\end{figure}  

\label{Spatial-Temporal Disstance}
It is worth noting that for a specific traffic graph $G$, the partitioning schemes $P1$ and $P2$ are predefined before the training phase and remain fixed during both training and inference. The application of these schemes is specifically structured within each STGAtt Block, which contains two sequential STGAtt Modules. As illustrated in Fig.\ref{fig:two_stage_partition}, these two modules work in tandem by cyclically applying the partition schemes: the first module partitions the data using $P1$ for its attention computation, and its output is then fed into the second module, which uses the shifted partition scheme $P2$ for its attention computation. This consecutive P1-then-P2 process within a single block ensures that state elements are regrouped into new neighborhood subsets, creating overlaps that act as bridges for information to flow between the otherwise isolated subsets from the first module. This entire sequence is repeated for every STGAtt Block stacked in the model, consistently facilitating information exchange across different spatial-temporal regions.

% It is worth noting that for a specific traffic graph $G$, the partitioning schemes $P1$ and $P2$ are predefined before the training phase and remain unchanged during both training and inference. Within the STGAtt model, we cyclically utilize these two partitioning schemes. As illustrated in Fig.\ref{fig:two_stage_partition}, assuming an initial partitioning scheme $P1$, after computing the attention, we first switch to $P2$ for the next attention computation and then revert back to $P1$, and so on.

\textbf{Spatial-Temporal Unified Graph Attention: }Given a state element neighborhood subset $\breve X$, we aim to dynamically assign different weights to the connections of each state element $\tilde X_{n,t} \in \breve X$ to adapt to their varying connection characteristics. The process begins at the single-head level with the $HeadAtt( \cdot )$ function. For a given target element $\tilde X_{n,t}$, it computes a weighted sum of all elements in its neighborhood $\breve X$, where each neighbor is first projected by a learnable matrix $W_1$ and then weighted by an attention score $\alpha$. Building upon this, the $Att^*( \cdot )$ function implements the multi-head mechanism for that single element. It computes $H$ parallel $HeadAtt()$ operations, concatenates their outputs, and applies a final linear transformation $W^H$ to produce an aggregated feature representation. Finally, the $Att( \cdot )$ function serves as a wrapper that applies this entire $Att^*( \cdot )$ process to every element within the input neighborhood subset $\breve X$, yielding a complete set of updated state representations. This process is as follows:

\begin{equation}
    \label{eq:multi_att}
    \begin{gathered}
        Att(\breve X) = \{Att^*(\tilde X_{n,t}) | \tilde X_{n,t} \in \breve X \} \hfill \\
        Att^*(\tilde X_{n,t}) = \left (\left\| {_{h = 1}^H} \right.HeadAtt{(\tilde X_{n,t}}) \right ){W^H} \hfill \\
        HeadAtt{(\tilde X_{n,t})} = \sum\nolimits_{\tilde X_{i,j} \in {{\breve X}}} {{\alpha _{((n,t),(i,j))}}} \cdot (\tilde X_{i,j}W_1) \hfill \\
    \end{gathered}
\end{equation}
Here, ${\alpha _{(( \cdot , \cdot),( \cdot , \cdot))}}$ represents the weight, i.e., attention score, assigned to the connection between two state elements, with $\sum \alpha = 1$. $W_1 \in {\mathbb{R}^{D \times \frac{D}{H}}},{W^H} \in {\mathbb{R}^{D \times D}}$ are learnable parameters, and $\left\| {_{h = 1}^H} \right. ( \cdot )$ denotes the concatenation of $H$ vectors. Eq.\ref{eq:multi_att} implements a multi-head attention mechanism \cite{vaswani2017attention} to enhance the learning capacity of the network. It is important to note that different attention heads have distinct learnable parameters.

Following the approach in \cite{vaswani2017attention}, we utilize inner product to calculate the weight score ${\alpha _{((n,t),(n',t'))}}$ between two connected state elements $\tilde X_{n,t}$ and $\tilde X_{n',t'}$. The weight scores are then normalized to ensure that the sum of all $\alpha$ equals 1.

\begin{equation}
    \begin{gathered}
        {\alpha _{((n,t),(n',t'))}} = \frac{{\exp ({\delta _{((n,t),(n',t'))}})}}{{\sum\nolimits_{\tilde X_{i,j} \in {{\breve X}}} {\exp ({\delta _{((n,t),(i,j))}})} }} \hfill \\
        {\delta _{((p,q),(p',q'))}} = \frac{{\left\langle {(\tilde X_{p,q}W_2 + b_2),(\tilde X_{p',q'}W_3 + b_3)} \right\rangle }}{{\sqrt \frac{D}{H} }} \hfill \\
    \end{gathered}
\end{equation}
where $\left\langle { \cdot , \cdot } \right\rangle $ represents the inner product between vectors, and $W_2,W_3 \in {\mathbb{R}^{D \times \frac{D}{H}}},b_2,b_3\in {\mathbb{R}^{ \frac{D}{H}}}$ are learnable parameters.

\textbf{Overall Architecture of STGAtt Block: }As depicted in Fig. \ref{fig:STGAtt_Architecture}, each STGAtt Block consists of two STGAtt Modules. Each module utilizes a different partitioning scheme ($P1$ or $P2$), with the target partition scheme set before data enters the module. 

Fig. \ref{fig:STGAtt_Architecture} illustrates the structure of an STGAtt Module, comprising a Spatial-Temporal Unified Graph Attention mechanism followed by a Feed-Forward Network (FFN) layer. Skip connections and normalization are applied within the module to enhance learning. Let $\tilde{X} \in \mathbb{R}^{N \times T \times D}$ denote the input to the STGAtt Module. The computational flow within the module is as follows:

% The STGAtt Module, as shown in Fig.\ref{fig:STGAtt_Architecture}, includes a Spatial-Temporal Unified Graph Attention mechanism followed by a Feed-Forward Networks (FFN) layer. Additionally, skip connections and normalization are applied within the module.  Let $\tilde{X} \in \mathbb{R}^{N \times T \times D}$ denote the input to the STGAtt Module. The computational flow within the module is as follows:

% 如图 \ref{fig:two_stage_partition} 中描述的那样，我们在一个 STGAtt Block 内部设置了两个 Spatial-Temporal Unified Graph Attention 计算模块，记为 STGAtt Module，并在计算的间隔重新划分状态分量邻域集合，以实现 Ontra-group 的信息交换。如图 \ref{fig:STGAtt_Architecture} 所示，在 STGAtt Module 中，除了 Spatial-Temporal Unified Graph Attention 外，还包含一个 Feed-Forward Networks (FFN) 层，并且skip connection and normalization are applied in our STGAtt Module。将STGAtt Module的输入记为 $\tilde Y \in {\mathbb{R}^{N \times T \times D}}$，STGAtt Module的计算流程如下：
\begin{equation}
    \begin{gathered}
      \tilde X = {Att}^{ST}(\tilde X)  + \tilde X \hfill \\
      \tilde X = Norm(\tilde X) \hfill \\
      \tilde X = FFN(\tilde X)  + \tilde X \hfill \\
      \tilde X = Norm(\tilde X) \hfill \\ 
    \end{gathered}
\end{equation}
Here, $Att^{ST}(\cdot)$ represents the result of applying Spatial-Temporal Unified Graph Attention to each state element neighborhood subset based on the predefined partitioning scheme using Eq.\ref{eq:multi_att}. The process can be elaborated as:

\begin{equation}
    \begin{gathered}
      \tilde X\xrightarrow{{Partition}} \{ {{\breve X}_{{i_1},{j_1}}},{{\breve X}_{{i_2},{j_2}}}, \ldots ,{{\breve X}_{{i_l},{j_l}}} \} \hfill \\
      \tilde X\xleftarrow{{Merge}} \{ Att({{\breve X}_{{i_1},{j_1}}}),Att({{\breve X}_{{i_2},{j_2}}}), \ldots ,Att({{\breve X}_{{i_l},{j_l}}}) \} \hfill \\ 
    \end{gathered}
\end{equation}
Initially, $\tilde{X}$ is divided into corresponding neighborhood subsets based on the predefined partitioning scheme. Spatial-Temporal Unified Graph Attention is then computed independently for each subset. Finally, the results from all subsets are merged to obtain the complete traffic flow embedding $\tilde{X}$. The FFN layer consists of two linear transformations with a $ReLU$ activation function:

\begin{equation}
    FFN(\tilde X) = ReLU(\tilde XW_1^f + b_1^f)W_2^f + b_2^f
\end{equation}
where $W_1^f \in \mathbb{R}^{D \times 4D}$, $W_2^f \in \mathbb{R}^{4D \times D}$, $b_1^f \in \mathbb{R}^{4D}$, and $b_2^f \in \mathbb{R}^D$ are learnable parameters.

% Additionally, as depicted in Fig.\ref{fig:STGAtt_Architecture}, a complete STGAtt Block is formed by performing a state element neighborhood subset partitioning operation between the two STGAtt Modules (i.e., between the two attention computation operations) and once at the end of the block.

\subsubsection{Dimensional Adapter}
To map the latent traffic flow representation $\tilde{X} \in \mathbb{R}^{N \times T \times D}$ to the data space and adapt the output of STGAtt to different prediction horizons, we employ a Dimensional Adapter module at the end of the network. This module generates the final prediction output $\hat{X} \in \mathbb{R}^{N \times T' \times C}$.

For each spatial node's state matrix $\tilde{X}_i^S \in \mathbb{R}^{T \times D}$ in $\tilde{X}$, we apply a linear transformation to adjust its temporal dimension to match the desired prediction horizon $T'$, resulting in the predicted traffic flow representation $\tilde{X}^{Pred} \in \mathbb{R}^{N \times T' \times D}$ in the latent space:

\begin{equation}
    \begin{gathered}
      \forall p \in \{ 1,2, \ldots ,N \}, \hfill \\
      \tilde X_{{S_p}}^{Pred} = W_1^{ada}\tilde X_p^S + b_1^{ada} \hfill \\ 
    \end{gathered}
\end{equation}
where $\tilde{X}_{{S_i}}^{Pred}$ represents the corresponding state matrix of the $i$-th spatial node in $\tilde{X}^{Pred}$, and $W_1^{ada} \in \mathbb{R}^{T' \times T}$, $b_1^{ada} \in \mathbb{R}^D$ are learnable parameters. Subsequently, we apply another linear transformation to transform $\tilde{X}^{Pred}$ into the final prediction output $\hat{X}$:

\begin{equation}
    \hat X = {{\tilde X}^{Pred}}W_2^{ada} + b_2^{ada}
\end{equation}
where $W_2^{ada} \in \mathbb{R}^{D \times C}$, $b_2^{ada} \in \mathbb{R}^C$ are learnable parameters.

\section{Experiments}

\subsection{Dataset}
% \begin{figure}
%   \centering
%   \begin{subfigure}[b]{0.45\textwidth}
%     \centering
%     \includegraphics[width=\textwidth]{Figures/pems-bay.jpg}
%     \caption{PEMS-BAY}
%     \label{fig:dataset:pems-bay}
%   \end{subfigure}
%   % \hfill
%   \begin{subfigure}[b]{0.45\textwidth}
%     \centering
%     \includegraphics[width=\textwidth]{Figures/sh-metro.png}
%     \caption{SHMetro}
%     \label{fig:dataset:sh-metro}
%   \end{subfigure}
%   \caption{Traffic graph of two datasets used in experiments.}
%   \label{fig:dataset}
% \end{figure}

\begin{table}[h]
    \centering
    \caption{
        The detial of datasets.
    }
    \begin{tabular}{c|c|c|c|c}
        \toprule
        \textbf{Dataset}  & \textbf{Nodes}  & \textbf{Edges}  & 
        \textbf{Time Interval}  & \textbf{Time Range}  \\
        \midrule
        PEMS-BAY  & 325  & 3887  & 5min   & 1/1/2017 - 3/31/2017 \\
        SHMetro   & 288  & 958   & 15min  & 7/1/2016 - 9/30/2016 \\
        \bottomrule
    \end{tabular}
    \label{tb:dataset}
\end{table}

To evaluate the performance of our STGAtt, we conducted traffic prediction experiments on two real-world traffic datasets: PEMS-BAY \cite{li2017diffusion} and SHMetro \cite{liu2020physical}. The PEMS-BAY dataset contains speed data collected from sensor nodes in the San Francisco Bay Area, where each node represents a traffic flow detector on a highway. The SHMetro dataset comprises in-flow and out-flow data collected from nodes in the Shanghai subway system, with each node representing a subway station. Table \ref{tb:dataset} summarizes the characteristics of both datasets. To ensure consistent temporal alignment and facilitate performance comparison, we used linear interpolation to resample the SHMetro data to a 5-minute time interval, matching that of the PEMS-BAY dataset.

We standardized both datasets using Z-Score normalization before feeding them into the model. For evaluation, following previous works, we partitioned the PEMS-BAY dataset into training, validation, and test sets with a 7:1:2 ratio. The SHMetro dataset was divided into training, validation, and test sets representing 62, 9, and 21 days of data, respectively.

\subsection{Parameter setting}
\begin{table}[]
    \centering
    \caption{Parameter settings used in our experiments}
    \begin{tabular}{c|c|c}
        \toprule
        \textbf{Parameters} & \textbf{PEMS-BAY} & \textbf{SHMetro} \\
        \midrule
        $T$     & 12    & 12    \\
        $H$     & 4     & 4     \\
        $l$     & 40    & 30    \\
        $R$     & 16    & 16    \\
        $D$     & 16    & 64    \\
        \midrule
        batch size      &   \multicolumn{2}{c}{8}      \\
        learning rate   &   \multicolumn{2}{c}{0.001}   \\
        epoch           &   \multicolumn{2}{c}{100}     \\
        \bottomrule
    \end{tabular} 
    \label{tab:setting}
\end{table}

Following previous works, we set the length of the input historical traffic flow observation data, denoted by $T$, to 12. This implies that traffic observations from the past 60 minutes are used for prediction. Our proposed prediction model employs a stack of $H$ STGAtt blocks and uses $R$ eigenvectors for spatial position encoding, with a feature embedding dimension $D$. The hyperparameters $H$, $R$, $D$, as well as the initial learning rate and batch size, were determined using Bayesian Search provided by the Microsoft Neural Network Intelligence (NNI) toolkit \cite{nni2021}. This optimization process resulted in setting the number of stacked STGAtt blocks $H=4$ and the number of eigenvectors $R=16$. The feature embedding dimension $D$ was set to 16 for the PEMS-BAY dataset and $D = 64$ for SHMetro.

\label{sec:base_node_selection}
The selection of base nodes and the threshold $\tau$ is critical for constructing the neighborhood subsets. Considering that the time dimension $T$ used in our experiments is much smaller than the number of spatial nodes $N$ (e.g., $T=12$ while $N=325$ for PEMS-BAY), we simplify the selection process to reduce complexity. Specifically, we confine the selection of base nodes to a single time step within the Spatial-Temporal Unified Graph ($G^{ST}$), choosing the central time step, $t_0 + \lfloor T/2 \rfloor$, as the source.

Our method proceeds as follows: First, we preset the number of neighborhood subsets, $l$, based on computational considerations. We then use $l$ as the number of clusters for the K-Means algorithm, which we apply to the spatial nodes at the central time step $t_0 + \lfloor T/2 \rfloor$. The nodes corresponding to the resulting cluster centroids are designated as the base nodes. With the base nodes fixed, we determine the value of $\tau$ iteratively. We start by incrementally increasing $\tau$ until the union of all neighborhood subsets defined by these base nodes first covers all state elements in the entire graph $G^{ST}$. The resulting set of base nodes and the corresponding $\tau$ are then finalized. Crucially, we enforce the constraint $\tau \ge \lfloor T/2 \rfloor$ throughout this process. This ensures that every neighborhood subset will contain state elements from each of the $T$ time steps, thus maintaining temporal completeness during the local attention computation. Following this procedure, we set $l=40$ for the PEMS-BAY dataset and $l=30$ for the SHMetro dataset.

The proposed model is trained using the Adam optimizer with the NNI-tuned initial learning rate of 0.001 and batch size of 8, for a total of 100 training epochs. Table \ref{tab:setting} summarizes the specific parameter settings. All experiments are conducted on a computer equipped with an Intel Xeon Gold 5218 CPU and a NVIDIA GeForce GTX 2080Ti GPU.

\subsection{Evaluation Metrics}
In order to verify performance of the proposed model for traffic flow prediction, the following three metrics are used for evaluation:

\begin{itemize}
    \item Mean Absolute Error (MAE): 
        \begin{equation}
            \begin{gathered}
                MAE(X,\hat X) = \frac{1}{{NT'}}\sum\nolimits_{n = 1}^N {\sum\nolimits_{t = 1}^{T'} {{{\left\| {{X_{n,t}} - {{\hat X}_{n,t}}} \right\|}_1}} } 
            \end{gathered}
        \end{equation}
        
    \item Mean Absolute Percentage Error (MAPE): 
        \begin{equation}
            \begin{gathered}
                MAPE(X,\hat X) = \frac{{100\% }}{{NT'}}\sum\nolimits_{n = 1}^N {\sum\nolimits_{t = 1}^{T'} {{{\left\| {\frac{{{X_{n,t}} - {{\hat X}_{n,t}}}}{{{X_{n,t}}}}} \right\|}_1}} } 
            \end{gathered}
        \end{equation}
        
    \item Root Mean Squared Error (RMSE): 
        \begin{equation}
            \begin{gathered}
                RMSE(X,\hat X) = \sqrt {\frac{1}{{NT'}}\sum\nolimits_{n = 1}^N {\sum\nolimits_{t = 1}^{T'} {{{({{\left\| {{X_{n,t}} - {{\hat X}_{n,t}}} \right\|}_2})}^2}} } } 
            \end{gathered}
        \end{equation}
\end{itemize}
where ${\left\|  \cdot  \right\|_1},{\left\|  \cdot  \right\|_2}$ are 1-norm and 2-norm respectively. It's worth noting that we exclude the prediction points where the true value is 0 from the calculation of the evaluation metrics. This is because traffic flow being 0 often indicates special circumstances like road closures or system errors. Including these points could disproportionately skew the evaluation results and misrepresent the model's performance under normal traffic conditions.
\section{Results}

\begin{table}[htb]
	\centering
            % The comparison of different approaches for traffic prediction. 最好的结果加粗显示，第二好的结果用下划线显示。Result show that our proposed model 在各方面都取得了最好的结果。
        \caption{
        Comparison of different approaches for traffic prediction. The best results are highlighted in \textbf{bold}, and the second-best results are \underline{underlined}. Results show that our proposed model achieves the best performance across the board. 
        }
        \resizebox{\textwidth}{!}{
            \begin{tabular}{@{}@{\extracolsep{\fill}}ll|ccc|ccc|ccc@{}}
                \toprule
                \multirow{2}{*}{Dataset}  & \multirow{2}{*}{Method}	& \multicolumn{3}{c}{15 min}	& \multicolumn{3}{c}{30 min}	& \multicolumn{3}{c}{1 hour}	\\
                &                 & MAE	& MAPE	& RMSE			& MAE	& MAPE	& RMSE			& MAE	& MAPE	& RMSE 			\\
                \midrule
                \multirow{9}{*}{\rotatebox{90}{PEMS-BAY}} 
                    & HA              & 2.88             & 6.85\%              & 5.59             & 2.88             & 6.85\%             & 5.59             & 2.88             & 6.85\%             & 5.59		        \\
                & FC-LSTM         & 2.05             & 4.80\%              & 4.19             & 2.20             & 5.20\%             & 4.55             & 2.37             & 5.70\%             & 4.96             \\
                & DCRNN           & 1.38             & 2.90\%              & 2.95             & 1.74             & 3.90\%             & 3.97             & 2.07             & 4.90\%             & 4.74             \\
                & STGCN           & 1.36             & 2.96\%              & 2.90             & 1.81             & 4.17\%             & 4.27             & 2.49             & 5.79\%             & 5.69             \\
                & GMAN            & 1.34             & 2.81\%              & 2.82             & 1.62             & 3.63\%             & 3.72             & 1.96             & 4.31\%             & 4.32             \\
                & STS-CCL         & 1.29             & 2.60\%              & 2.54             & 1.61             & 3.70\%             & 3.96             & 1.90             & 4.24\%             & 4.07             \\
                & STSGCN          & \underline{1.25} & \underline{2.54\% } & \underline{2.43} & \underline{1.57} & \underline{3.34\%} & \underline{3.30} & 1.92             & 4.35\%             & 4.10		        \\
                & ASTTN           & 1.32             & 2.78\%              & 2.70             & 1.58             & 3.64\%             & 3.72             & \underline{1.72} & \underline{3.98\%} & \underline{4.02} \\
                    & STGAtt (Ours)   & \textbf{1.13}    & \textbf{2.30\%}     & \textbf{2.33}    & \textbf{1.38}    & \textbf{3.02\%}    & \textbf{3.06}    & \textbf{1.67}    & \textbf{3.78\%}    & \textbf{3.80}    \\
                \midrule	
                \multirow{9}{*}{\rotatebox{90}{SHMetro}} 	
                    & HA              & 48.26             & 31.55\%             & 136.97            & 47.88             & 31.49\%             & 136.81            & 46.40             & 30.80\%             & 135.72		    \\
                & FC-LSTM         & 26.68             & 18.76\%             & 55.53             & 27.25             & 19.04\%             & 57.37             & 28.94             & 20.59\%             & 63.41             \\
                & DCRNN           & 24.04             & 17.82\%             & 46.02             & 25.23             & 18.35\%             & 49.90             & 28.01             & 20.44\%             & 58.83             \\
                & STGCN           & 24.70             & \underline{16.91\%} & 48.03             & 25.02             & 17.05\%             & 48.99             & 27.52             & 17.50\%             & 55.82             \\
                & GMAN            & 24.52             & 17.45\%             & 53.06             & 26.44             & 20.47\%             & 61.43             & 29.42             & 21.40\%             & 70.11             \\
                & STS-CCL         & 23.65             & 17.33\%             & 45.51             & 24.52             & 17.64\%             & 47.74             & 25.49             & 17.36\%             & 51.16             \\
                & STSGCN          & \underline{23.16} & 17.60\%             & \underline{43.86} & \underline{23.44} & 17.48               & \underline{44.35} & \underline{24.36} & 17.80\%             & 46.19             \\
                & ASTTN           & 24.23             & 18.01\%             & 46.98             & 23.78             & \underline{17.26\%} & 45.01             & 24.86             & \underline{17.13\%} & \underline{45.74} \\
                    & STGAtt (Ours)   & \textbf{22.19}    & \textbf{16.64\%}    & \textbf{41.69}    & \textbf{22.72}    & \textbf{16.79\%}    & \textbf{43.14}    & \textbf{23.73}    & \textbf{17.07\%}    & \textbf{45.62}    \\	
                \bottomrule
            \end{tabular}
        }
        \label{tb:baseline_result}
\end{table}

To evaluate the competitive performance of our proposed model (STGAtt), we compare it against the following optimized baselines:

\begin{itemize}
    \item HA (Historical Average): This baseline treats traffic states as strictly periodic, predicting the current state using the average of corresponding states from several previous periods. In this study, we use a weekly period, averaging traffic states at the same time point across multiple preceding weeks.

    \item FC-LSTM (Fully-Connected Long Short-Term Memory Network) \cite{sutskever2014sequence}: This baseline employs an encoder-decoder framework with fully connected LSTM units.

    \item DCRNN (Diffusion Convolutional Recurrent Neural Network) \cite{li2017diffusion}: This model integrates diffusion graph convolution networks with recurrent neural networks within an encoder-decoder framework.

    \item STGCN (Spatio-Temporal Graph Convolutional Network) \cite{yu2017spatio}: This approach leverages graph convolution to capture spatial dependencies and gated 1D convolution to capture temporal dependencies.

    \item GMAN (Graph Multi-Attention Network) \cite{zheng2020gman}: This model employs two independent sets of multi-head attention modules to capture both temporal and spatial dependencies.

    \item STS-CCL (Spatial-Temporal Synchronous Contextual Contrastive Learning (STS-CCL) model) \cite{li2024sts}: This model utilizes an Informer\cite{zhou2020informer}-based architecture (an attention-based model) for the temporal dimension and GCN for the spatial dimension.

    \item STSGCN (Spatio-Temporal Synchronous Graph Convolutional Network) \cite{song2020spatial}: This approach expands the spatial network topology across the temporal dimension to create a spatial-temporal network. It then uniformly applies graph convolution to capture synchronous spatial and temporal correlations.
        
    \item ASTTN (Adaptive Graph Spatial-Temporal Transformer Network) \cite{feng2022adaptive}: Similar to STSGCN, ASTTN constructs a cross-dimensional network. However, it solely relies on an attention mechanism.
\end{itemize}

In summary, our proposed method and the baseline approaches can be categorized as follows:

% \begin{itemize}
%     \item Traditional Statistical Method: HA.

%     \item Temporal Modeling: FC-LSTM \cite{sutskever2014sequence}.

%     \item Separate Spatial and Temporal Modeling: DCRNN (GCN, RNN) \cite{li2017diffusion}, STGCN (GCN, CNN) \cite{yu2017spatio}, GMAN (Attention, Attention) \cite{zheng2020gman}, STS-CCL (GCN, Attention) \cite{li2024sts}.  (Parentheses indicate the methods used for spatial and temporal modeling, respectively).

%     \item Unified spatial-temporal Modeling: STSGCN (GCN) \cite{song2020spatial}, ASTTN (Attention) \cite{feng2022adaptive}, Ours (Graph-Based Attention). These approaches construct a unified graph representation across spatial and temporal dimensions to capture spatial-temporal dependencies.
% \end{itemize}

\begin{itemize}
    \item Traditional Statistical Method: HA.

    \item Temporal Modeling: FC-LSTM.

    \item Separate Spatial and Temporal Modeling: DCRNN (GCN, RNN), STGCN (GCN, CNN), GMAN (Attention, Attention), STS-CCL (GCN, Attention).  (Parentheses indicate the methods used for spatial and temporal modeling, respectively).

    \item Unified spatial-temporal Modeling: STSGCN (GCN), ASTTN (Attention), Ours (Attention). These approaches construct a unified graph representation across spatial and temporal dimensions to capture spatial-temporal dependencies.
\end{itemize}

Table \ref{tb:baseline_result} presents the results of traffic prediction experiments conducted on the PEMS-BAY and SHMetro datasets. Model performance was evaluated across three prediction horizons: 15 minutes, 30 minutes, and 60 minutes. Our proposed STGAtt model consistently outperforms other baseline models across all prediction horizons and evaluation metrics (MAE, MAPE, and RMSE). This superior performance highlights the effectiveness of our attention mechanism based on the Spatial-Temporal Unified Graph, coupled with the state element partitioning and exchanging mechanism, in capturing both short-range and long-range spatial-temporal dependencies. Specifically, we observe the following:
\begin{enumerate}
    \item \textbf{Performance Variation of HA: }The significant performance difference of the statistical method HA across the two datasets suggests distinct spatial-temporal characteristics of the PEMS-BAY and SHMetro datasets. The consistently strong performance of our STGAtt model on both datasets underscores its robust forecast ability.
    
    % 统计学方法HA的效果在两个数据集中差距显著，说明个两个实验数据集：PEMS-BAY，SHMetro 在时空特性上的显著差异。我们的STGAtt可以在两个数据集上都取得最好的结果，说明了我们的STGAtt良好的泛化能力。

    \item \textbf{Importance of Spatial and Temporal Modeling: }The FC-LSTM model, incorporating temporal modeling, shows a clear improvement over the HA baseline. Methods that model spatial dependencies (DCRNN, STGCN, GMAN, STS-CCL) further enhance performance, emphasizing the importance of capturing both spatial and temporal dependencies in traffic flow prediction. The superior results achieved by methods employing unified spatial-temporal modeling (STSGCN, ASTTN, and our STGAtt) further demonstrate the significance of capturing cross-dimensional relationships directly. Our STGAtt's ability to effectively differentiate between connections across spatial and temporal dimensions allows it to capture intricate spatial-temporal correlations within traffic flow signals, leading to its superior performance. 
    
    % Temporal Modeling 的 FC-LSTM 相比 HA 获得了明显的提升，而 model spatial 的方法：DCRNN, STGCN, GMAN, STS-CCL等，由进一步地提升了性能。无疑揭示了 Spatial and Temporal 依赖的捕获，在交通流预测中的重要性。再观察 Unified spatial-temporal Modeling 的方法，它们在 traffic prediction 的实验中取得了最好的结果，继续证明了跨越 spatial 和 temporal 维度关系的重要性。我们的 STGAtt 在其中取得了最好的性能，说明了 STGAtt 成功地处理了跨维度连接关系区别，捕获了交通流信号中的各种复杂的时空关系。

    \item \textbf{Strengths in Short-range and Long-range Dependency Capturing: }Analyzing the second-best performing methods on both datasets reveals that within the unified spatial-temporal modeling category, the GCN-based STSGCN excels in short-term prediction (15min, 30min), while the attention-based ASTTN performs better in long-term prediction (60min). This suggests the significance of considering physical graph structure for shorter prediction horizons and the strength of attention mechanisms in capturing long-range dependencies. Our STGAtt model demonstrates superior performance across both short and long prediction horizons for the following reasons:
        \begin{itemize}
            \item \textbf{Short-Term Performance:} By computing attention within state element neighborhood subsets, STGAtt is able to effectively focus on short-range spatial-temporal relationships, akin to the approach used in GCNs, leading to its success in short-term prediction.
        
            \item \textbf{Long-Term Performance:} The exchange of state elements between subsets effectively facilitates inter-subset information interaction, allowing STGAtt to capture global spatial-temporal information effectively. This capability enables the model to handle long-range spatial-temporal dependencies successfully.
        \end{itemize}
    In conclusion, the combined partitioning and exchange strategy allows our STGAtt model to effectively capture both short-range and long-range spatial-temporal dependencies, ultimately leading to its exceptional prediction performance. 
    
    % \item 进一步观察两个数据集中第二好的方法，可以发现，在Unified spatial-temporal Modeling 的方法中，短时间预测（15min，30min）中，基于图卷积的STSGCN方法占据优势；而长时间预测（60min），基于注意力的方法ASTTN，则效果更好。表明了考虑物理图结构在较短时长的预测中的重要作用，而注意力模型能更擅长关于长距离的依赖信息。而我们的 STGAtt 可以在短时间和长时间内都取得最好的效果：
    %     \begin{itemize}
    %         \item 说明我们在状态分量邻域子集内部计算注意力的方法，让 STGAtt 成功考虑到了物理图结构的信息，使其能如GCN一样，成功关注于短距离的时空关系，从而在短时间预测取得成功。

    %         \item 长时间预测的成功，说明了子集之间交换状态分量的方法成功地实现了子集间的信息交互，让 STGAtt 能更高效地捕获到全局的时空信息，从而能顺利处理长距离的时空关系。
    %     \end{itemize}
    % 总的来说，划分与交换结合的方案成功地让我们的 STGAtt 能同时胜任长距离和短距离的时空关系的捕获，最终获得优异的预测性能。
\end{enumerate}
\section{Discussion}

\begin{figure}[ht]
    % \begin{adjustwidth}{-\extralength}{0cm}
    \centering
    
    \subfloat[Sunday 07:00 - 07:55\centering]{
        \includegraphics[width=0.45\textwidth]{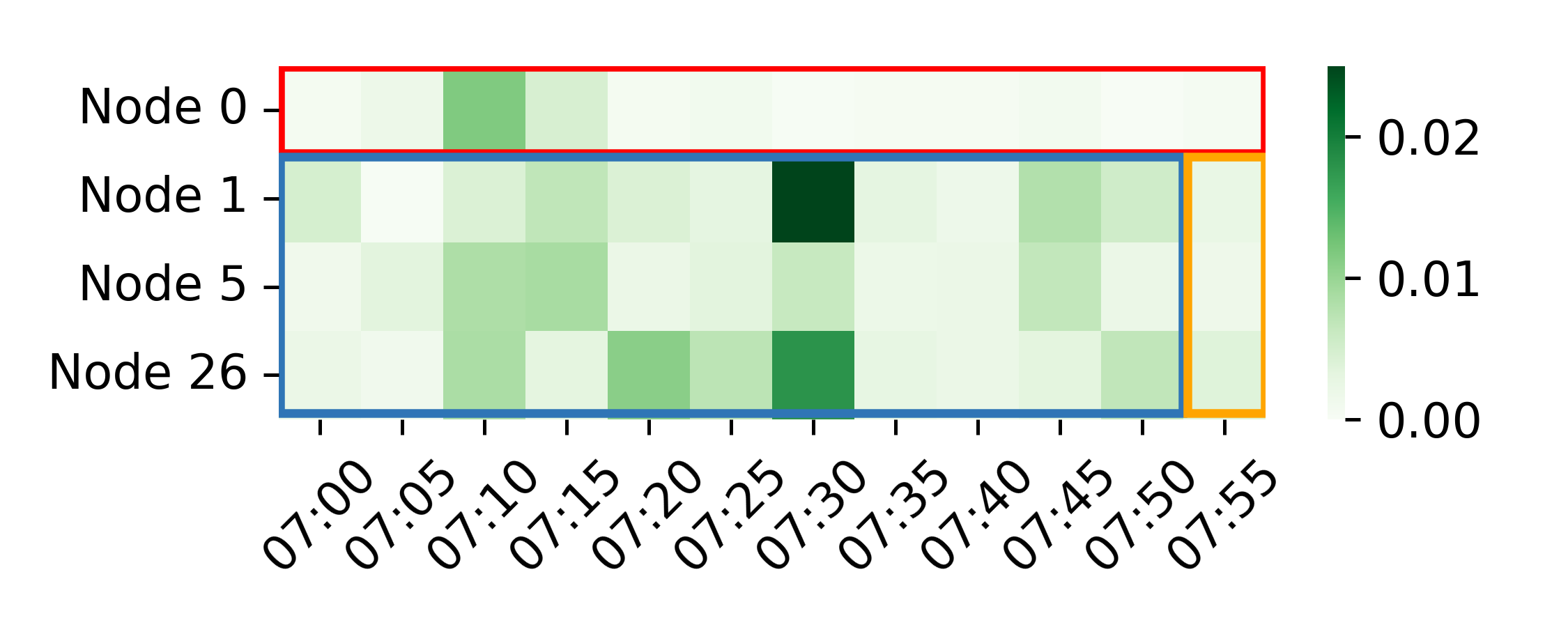}
    }
    \subfloat[Monday 07:00 - 07:55\centering]{
        \includegraphics[width=0.45\textwidth]{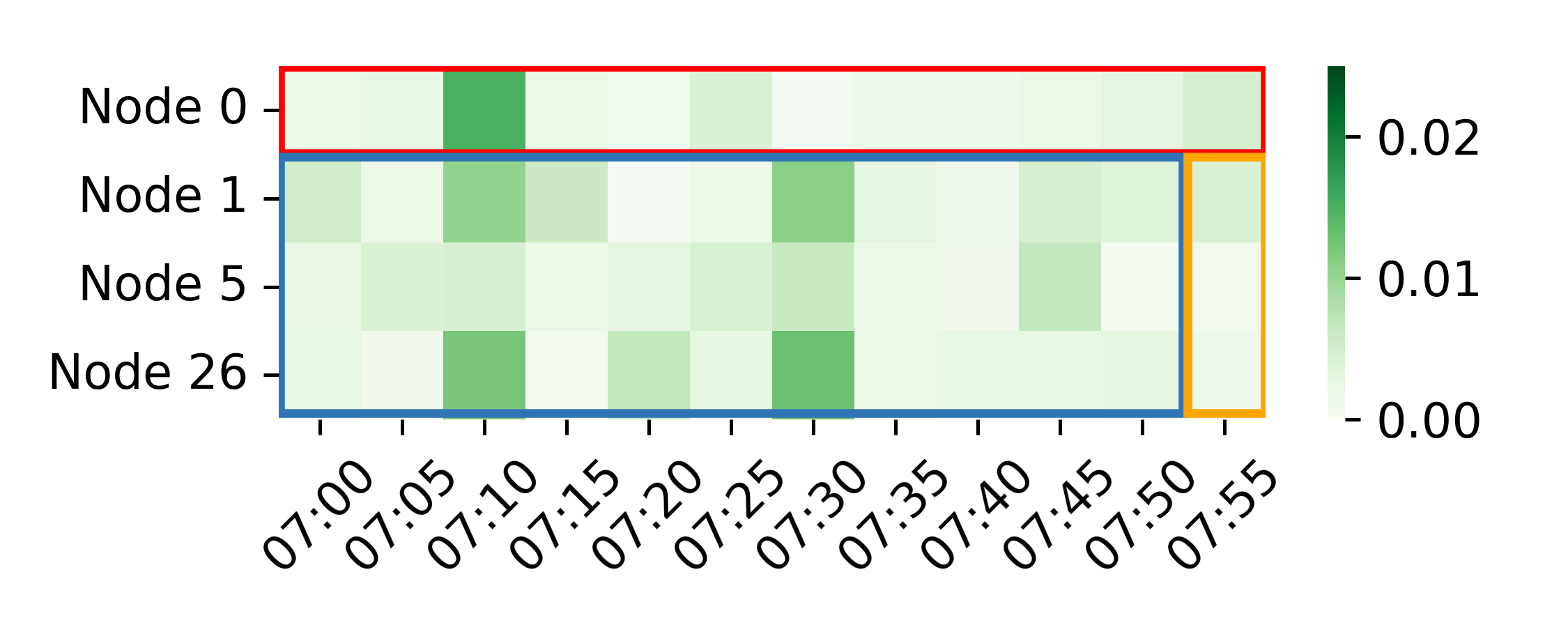}
    } \\
    
    \subfloat[Wednesday 12:00 - 12:55\centering]{
        \includegraphics[width=0.45\textwidth]{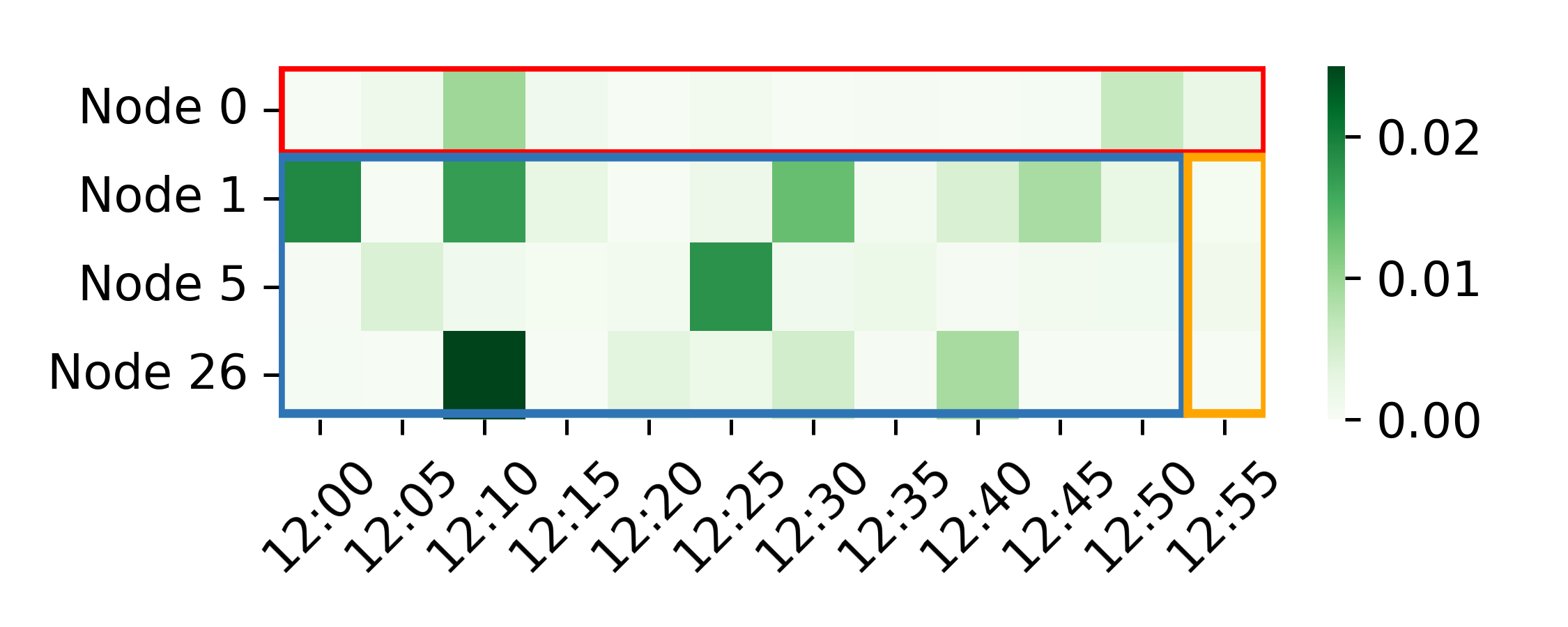}
    }
    \subfloat[Friday 18:00 - 18:55\centering]{
        \includegraphics[width=0.45\textwidth]{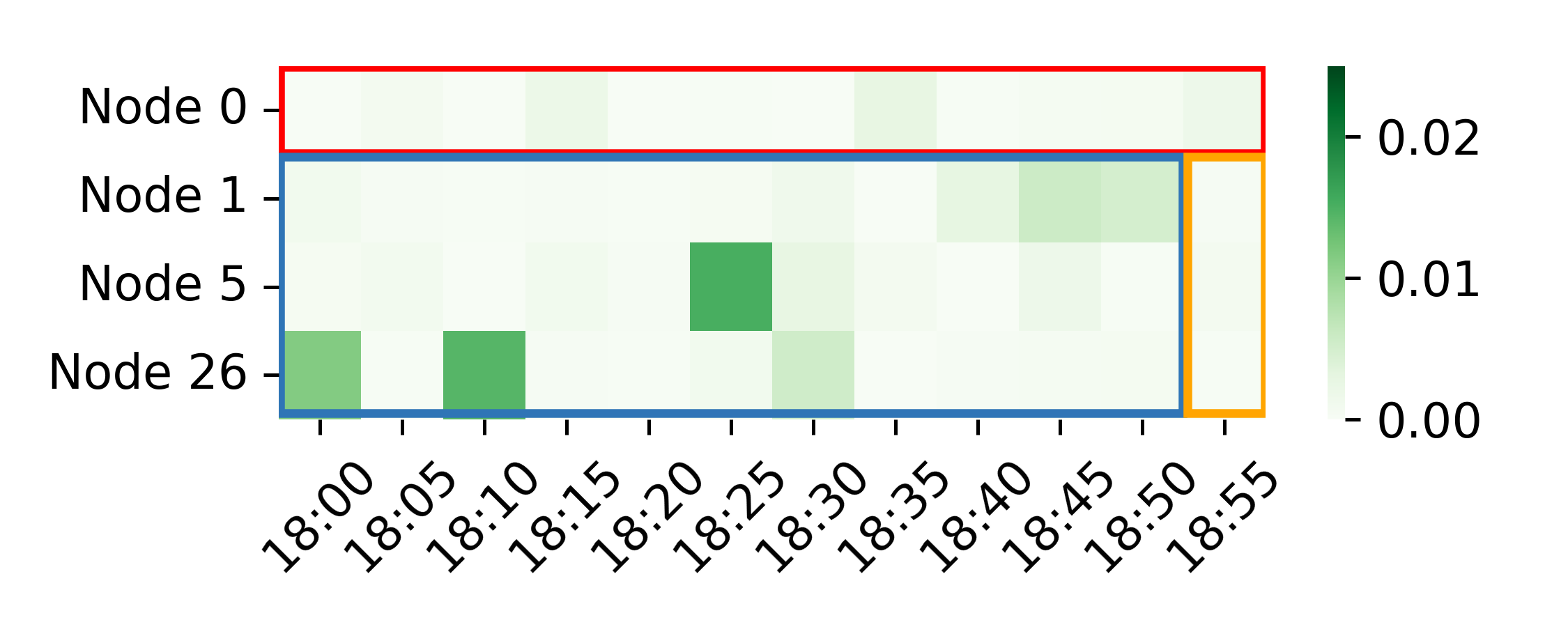}
    } \\
    
    % PEMS-BAY 数据集 Node 0, Node 1, Node 5 and Node 26 一个星期的速度数值。灰色竖直虚线标记了每天的开始与结束。
    \subfloat[Weekly speed values for node 0, node 1, node 5, and node 26 from the PEMS-BAY dataset. The grey vertical dashed lines mark the beginning and end of each day.\centering]{
        \includegraphics[width=0.9\textwidth]{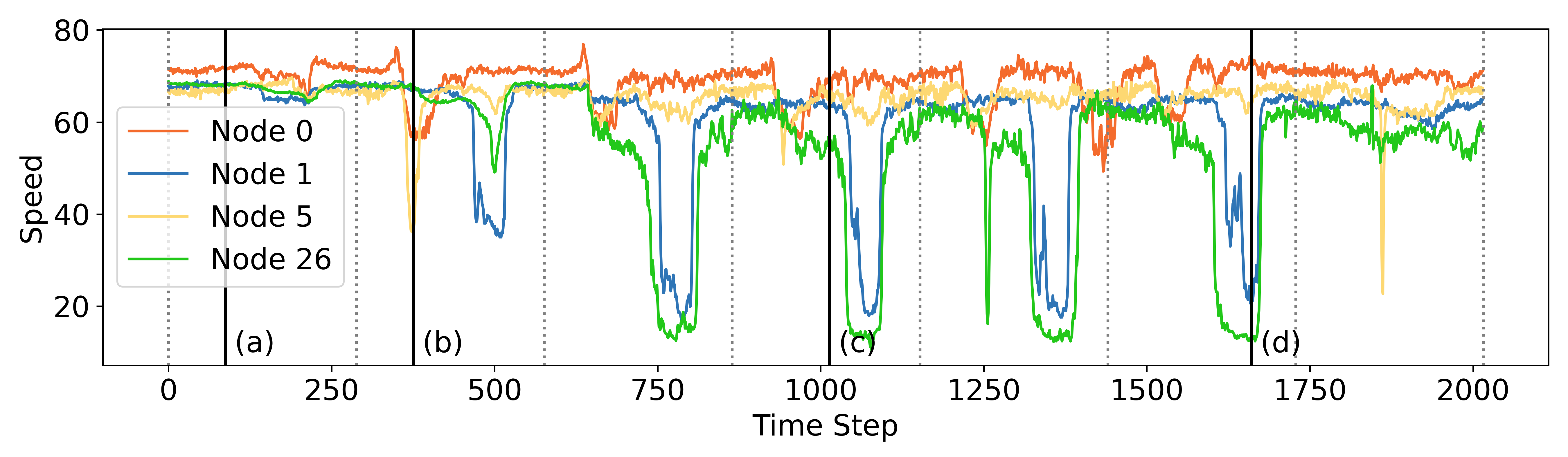}
    }
    \caption{
        Attention score heatmaps (a-d) for selected nodes in the PEMS-BAY dataset, juxtaposed with their corresponding speed values (e). The red, orange, and blue boxes within the heatmaps represent temporal, spatial, and spatial-temporal influences, respectively. The vertical solid lines in the speed plot (e) correspond to the specific time snapshots depicted in (a-d). As evident, traffic patterns change significantly at different times. Our STGAtt model effectively captures these dynamic fluctuations and their influence on traffic flow.
    }
    \label{fig:pems_vis_att}
\end{figure}

To validate the effectiveness of our STGAtt model, we visualize the attention scores, represented by  ${\alpha _{(( \cdot , \cdot),( \cdot , \cdot))}}$ in Eq.\ref{eq:multi_att}, for several prediction samples from both the PEMS-BAY and SHMetro datasets. The attention score between two state elements in a traffic flow signal reflects their correlation, which is inherently governed by their spatial-temporal relationships.

Fig.\ref{fig:pems_vis_att}(a-d) presents the attention scores between corresponding state elements of node 0, node 1, node 5, and node 26 in the PEMS-BAY dataset. The time range for each subfigure is indicated in its title. The vertical axis of each subfigure represents the node $v$ corresponding to the state element, while the horizontal axis represents the time $t$. The value in the heatmap represents the attention score ${\alpha _{(( v , t),( \text{Node 0} , t_{max}))}}$ between the state element ${{\tilde X}_{v , t}}$ and the state element ${{\tilde X}_{0 , t_{max}}}$ of node 0 at the latest time ($t_{max}$, the rightmost point on the horizontal axis) in that subfigure. The attention subfigures demonstrate three distinct influence relationships: temporal (red box), spatial (orange box), and spatial-temporal (blue box). Taking Fig.\ref{fig:pems_vis_att}a ($t_{max}=07:55$) as an example:

\begin{itemize}
    \item The cell at row "Node 0" (1st row) and column "07:20" (5th column) represents ${\alpha _{(( \text{Node 0} , \text{07:20}),( \text{Node 0} , \text{07:55}))}}$. Similarly, the attention scores in this row only differ in time, illustrating the temporal influence.

    \item The cell at row "Node 5" (3rd row) and column "07:55" (12th column) represents ${\alpha _{(( \text{Node 5} , \text{07:55}),( \text{Node 0} , \text{07:55}))}}$. The attention scores in this column only differ in nodes, demonstrating the spatial influence.

    \item The cell at row "Node 26" (4th row) and column "07:25" (6th column) represents ${\alpha _{(( \text{Node 26} , \text{07:25}),( \text{Node 0} , \text{07:55}))}}$. The attention scores in this region differ in both node and time, showcasing the spatial-temporal influence.
    
\end{itemize}
Fig.\ref{fig:pems_vis_att}e shows the weekly speed variations for node 0, node 1, node 5, and node 26. The vertical dashed lines mark the time points corresponding to the four subfigures (a-d). We have chosen four time periods with distinct characteristics, each corresponding to a subfigure:

\begin{itemize}
    \item Subfigure a: Sunday 07:00 - 07:55. During this non-working day period, the speeds of the four nodes are close to free-flow speed, indicating low traffic volume.

    \item Subfigure b: Monday 07:00 - 07:55. This corresponds to the same time period as subfigure a but on a working day. The speed plot reveals that this is the morning peak hour, where the speeds of node 0 and node 5 have decreased, suggesting increased traffic flow; while node 1 and node 26 show little change.

    \item Subfigure c: Wednesday 12:00 - 12:55. During this midday period on a working day, most nodes' speeds are close to free-flow speed, except for node 26, which shows a decrease in speed.

    \item Subfigure d: Friday 18:00 - 18:55. During this evening peak hour on a working day, the speeds of node 1 and node 26 drop significantly, indicating heavy traffic flow; while node 0 and node 5 remain relatively unchanged.

\end{itemize}
Observing the attention heatmaps, we can make the following observations:

\begin{enumerate}
    \item Spatial-temporal influence is dominant over purely temporal or spatial influences. In all four subfigures of Fig. \ref{fig:pems_vis_att}, the purely temporal correlations (the red-boxed row for Node 0) and purely spatial correlations (the orange-boxed column at time $t_{max}$) consistently show lighter colors, indicating lower attention scores. In contrast, the highest attention scores (the darkest green cells) are almost always found within the spatial-temporal domain (the blue-boxed area).

    \item The learned spatial-temporal correlations are highly dynamic and adaptive to different traffic scenarios. The model dynamically adjusts which nodes and time steps are most important based on the current traffic conditions. For example: During the Monday morning peak (Fig. \ref{fig:pems_vis_att}b), when speeds at Node 0 and Node 5 are decreasing (see Fig. \ref{fig:pems_vis_att}e), the model pays higher attention to Node 1 and Node 26. In contrast, during the Friday evening peak (Fig. \ref{fig:pems_vis_att}d), the traffic pattern shifts, with significant speed drops at Node 1 and Node 26. The model's attention adapts accordingly, now focusing heavily on Node 5 and Node 26 to predict the state of Node 0. This demonstrates a key advantage over traditional convolution-based models with fixed receptive fields; our STGAtt model can learn and adapt to these dynamically changing correlations, capturing the nuanced realities of traffic flow.

    \item The model holistically captures complex cross-dimensional relationships. The visualizations confirm that our STGAtt model does not simply treat spatial and temporal features as separate entities. Instead, by identifying the most influential nodes at specific past time steps for a given prediction (as evidenced by the shifting hotspots across Fig. \ref{fig:pems_vis_att}(a-d)), the model effectively captures the underlying, integrated spatial-temporal structure of traffic data. This unified modeling approach is what empowers STGAtt to achieve more accurate traffic flow predictions, particularly in complex and dynamic scenarios like rush hours.
\end{enumerate}

\begin{figure}[H]
        % \begin{adjustwidth}{-\extralength}{0cm}
    \centering
    
    \subfloat[Friday 06:00\centering]{
        \includegraphics[width=0.45\textwidth]{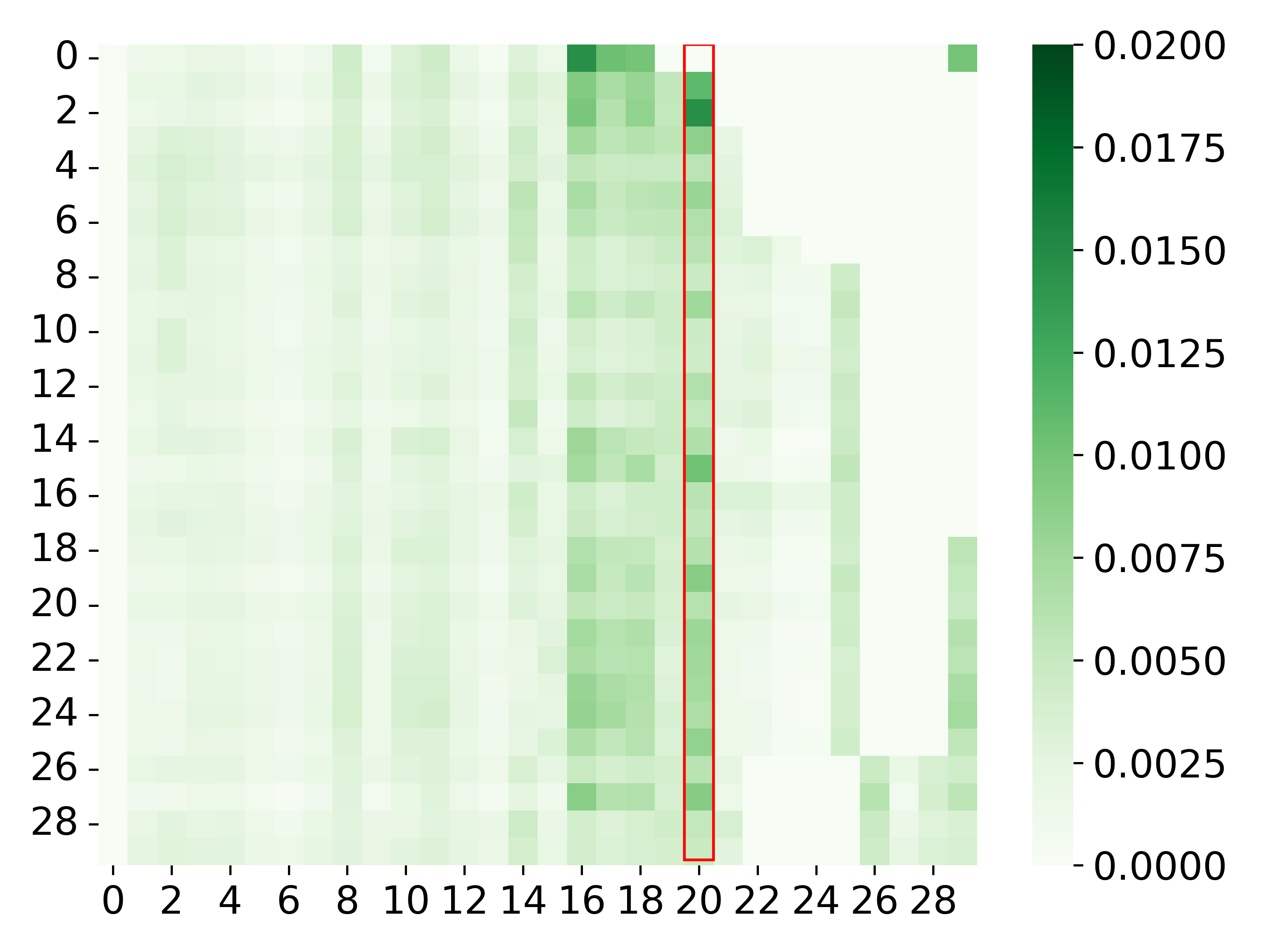}
    }
    \subfloat[Friday 09:00\centering]{
        \includegraphics[width=0.45\textwidth]{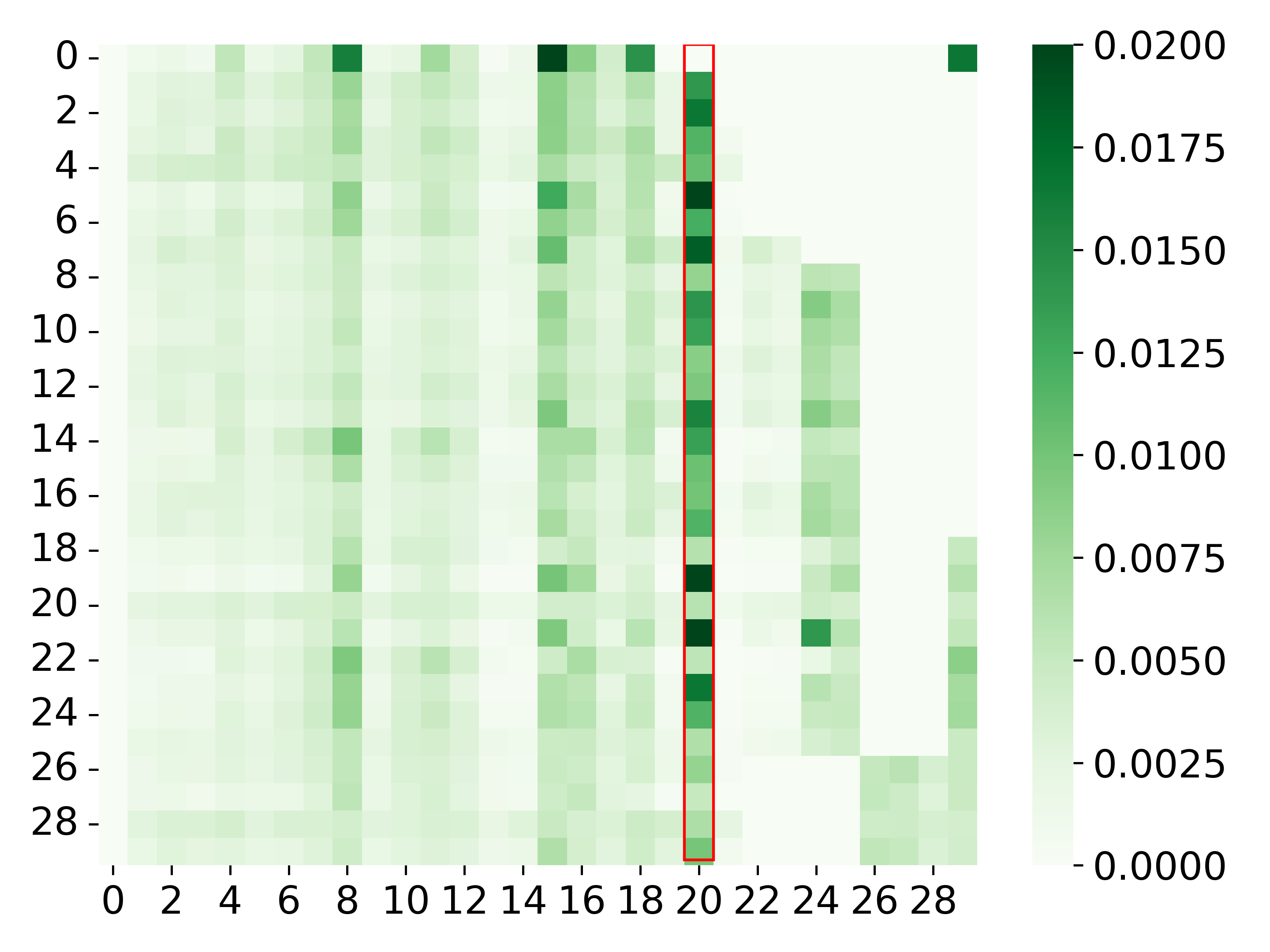}
    } \\
    
    \subfloat[Friday 12:00\centering]{
        \includegraphics[width=0.45\textwidth]{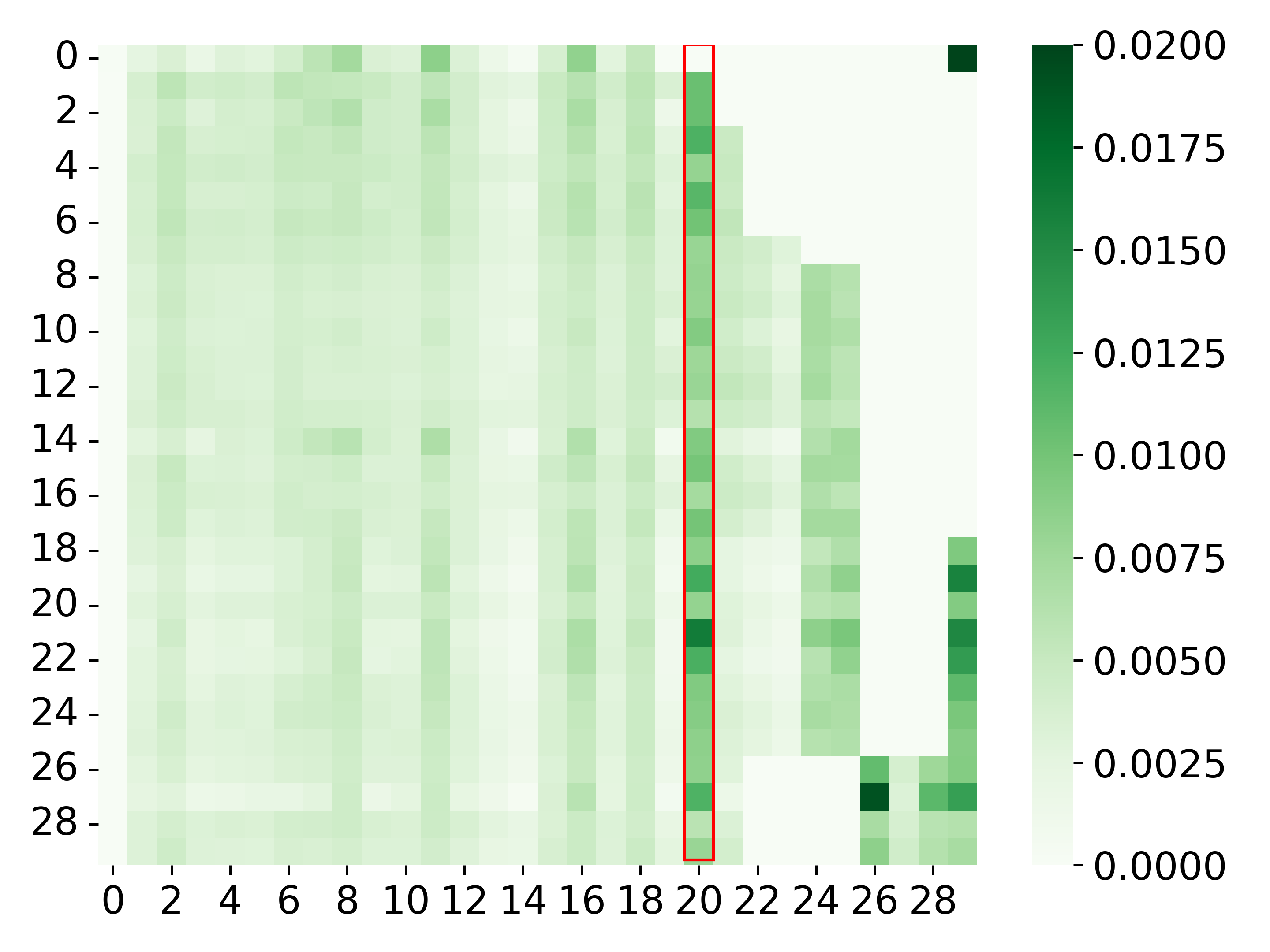}
    }
    \subfloat[Friday 15:00\centering]{
        \includegraphics[width=0.45\textwidth]{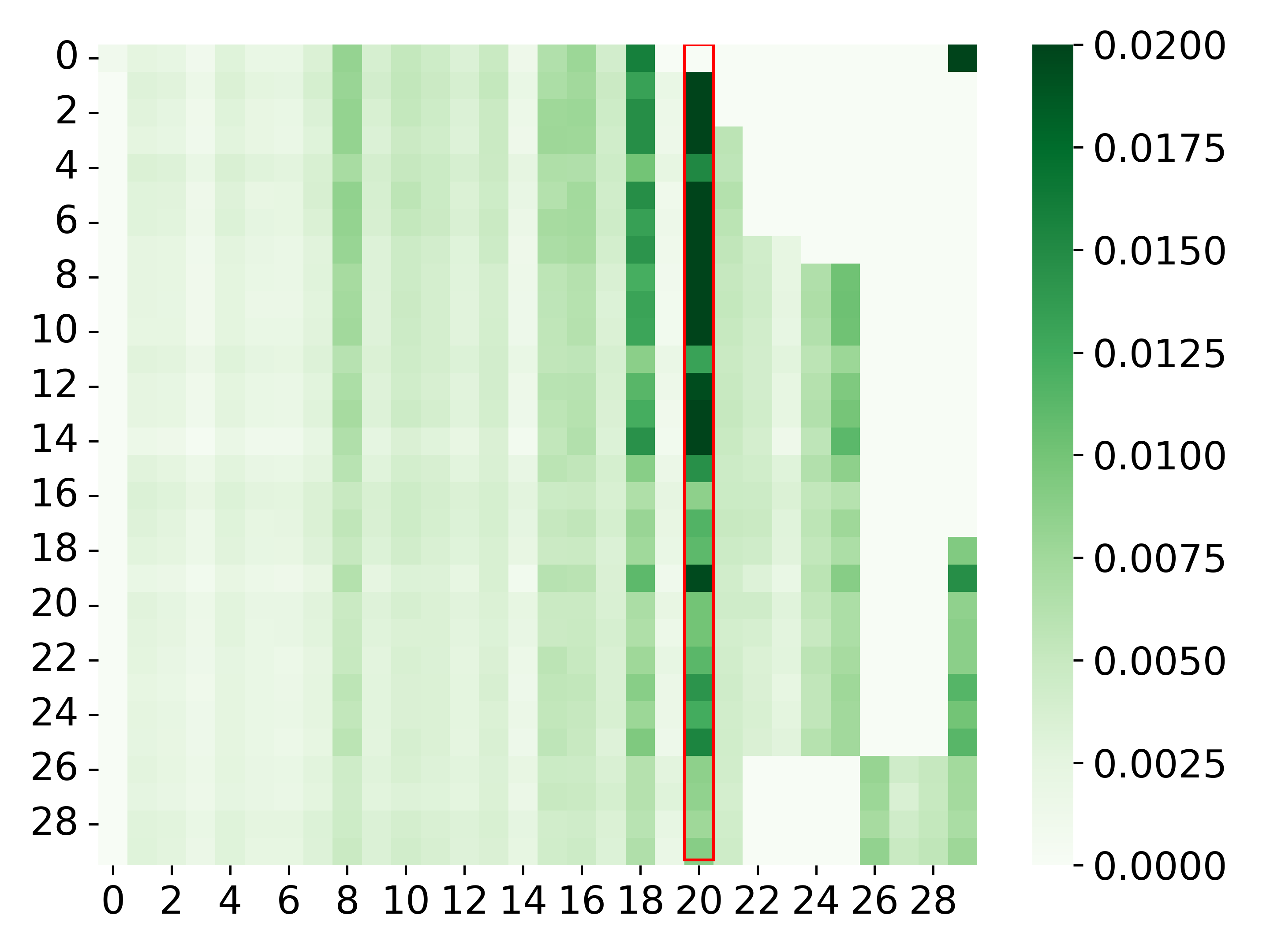}
    } \\
    
    % SHMetro 数据集 Node 20 所处位置 （深色标记），以及周边的节点（浅色标记）。
    \subfloat[Relative location of node 20 (dark marker) within the SHMetro dataset, along with its neighboring nodes (light markers). \centering]{
        \includegraphics[width=0.7\textwidth]{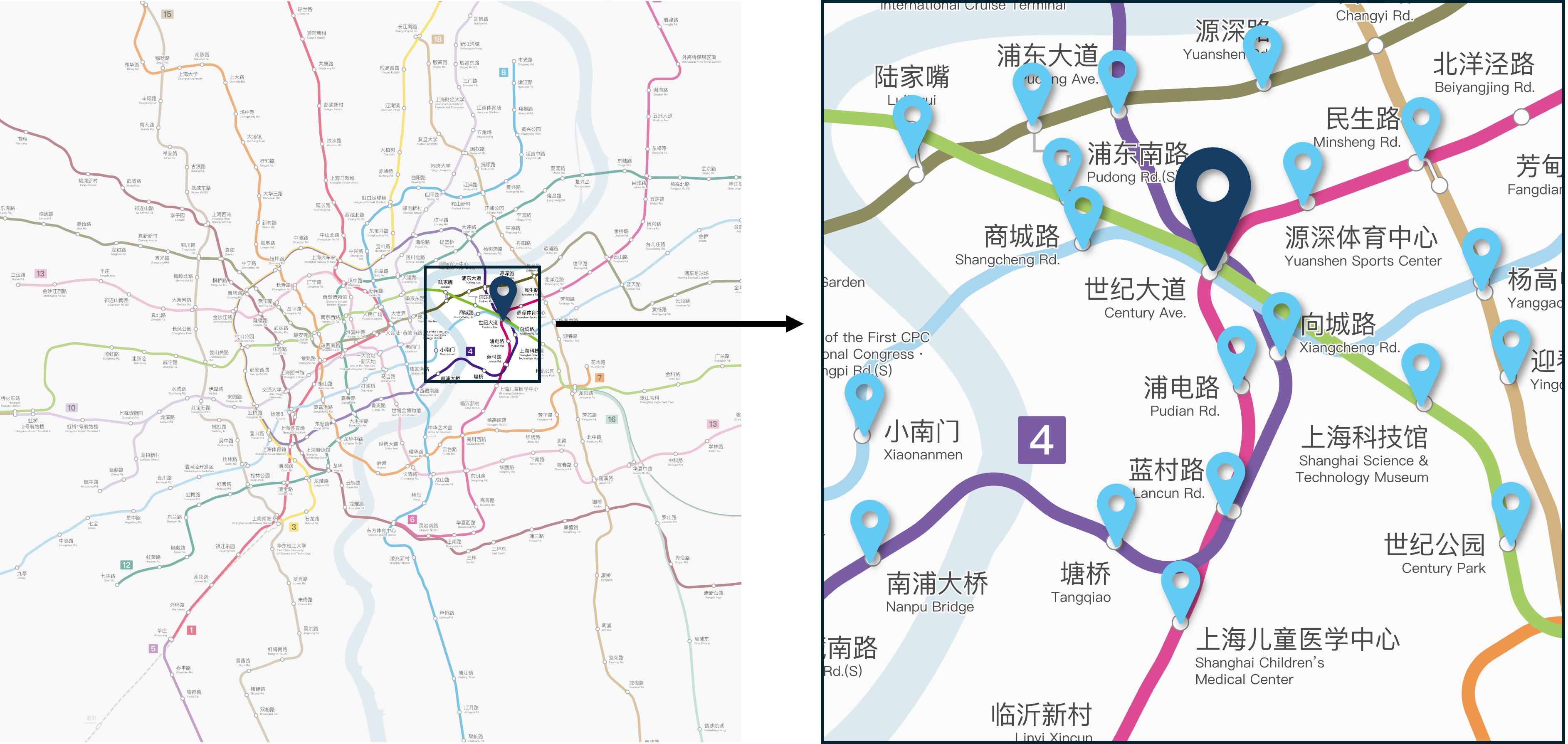}
    }
    % \end{adjustwidth}
    \caption{
        Attention score heatmaps for selected nodes in the SHMetro dataset, presented alongside their relative positions (e). Notably, our STGAtt model successfully identifies the far-reaching influence of the central node (node 20) on its surrounding nodes within the road network.
        % SHMetro 数据集部分节点的注意力分数热力图 (a-d)，以及对应的相对位置 (e)。可以看出，我们的 STGAtt 可以成功捕捉到路网里面中心节点对周围节点的大范围影响。
    }
    \label{fig:sh_vis_att}
\end{figure}

Next, we visualize the attention scores between nodes within a state element neighborhood subset in the SHMetro dataset. As shown in Fig.\ref{fig:pems_vis_att} and discussed previously, Spatial-temporal influence is dominant in the traffic flow signal. Therefore, in Fig.\ref{fig:sh_vis_att}, we focus on visualizing the attention scores representing this influence. Specifically, the vertical axis in the subfigure (a-d) of Fig.\ref{fig:sh_vis_att} represents the nodes corresponding to the state elements at time $t$ (indicated in the subfigure titles), while the horizontal axis represents the nodes at time $t-30min$. Thus, subfigure (a-d) illustrate the pairwise attention scores between nodes within the state element neighborhood subset, showcasing the spatial-temporal influence. Fig.\ref{fig:sh_vis_att}(e) shows the relative positions of these nodes and their physical connections, highlighting node 20, the central node with the highest degree.

The subfigures (a-d) of Fig.\ref{fig:sh_vis_att} depict four different time periods on Friday, each with distinct traffic characteristics. Observing the heatmaps corresponding to these four time periods, we can identify a dark vertical band at the location of node 20, marked by red boxes, indicating the strong influence of the neighboring nodes on this node. The prominent vertical band at node 20 is consistently observed across all four distinct time periods shown in subfigures (a-d), from the early morning commute to the mid-afternoon. This signifies that node 20 is a persistently influential node, affecting the future state of numerous other nodes across the graph regardless of the time of day. The persistence of this influence strongly suggests it is rooted in the node's intrinsic topological importance, rather than a temporary traffic pattern. This is confirmed by its strategic position within the transit network: Fig. \ref{fig:sh_vis_att}(e) identifies node 20 as Century Avenue station, a critical interchange connecting multiple metro lines. Such hubs act as major sources and sinks of traffic flow, and their conditions propagate widely throughout the day. STGAtt's ability to capture this persistent, long-range influence is a testament to its effectiveness. Instead of only learning from immediate neighbors, the model successfully identifies functionally critical nodes and their stable, long-range dependencies, demonstrating a deeper, global understanding of the network's dynamics. This global awareness is a key contributor to our model's superior predictive performance.

\section{Conclusion}

This paper introduces STGAtt, a novel deep learning model for traffic flow forecasting that excels at capturing complex spatial-temporal dependencies. STGAtt leverages a unified graph representation and an attention mechanism to dynamically weigh spatial and temporal connections, overcoming limitations of traditional approaches that model these dependencies separately.  By further partitioning state elements into localized subsets and employing a novel exchanging mechanism, STGAtt effectively captures both short-range and long-range correlations.

Extensive experiments on two real-world datasets demonstrate STGAtt's superior performance compared to state-of-the-art baselines. Visualizations of attention weights highlight its ability to adapt to dynamic traffic patterns and learn long-range dependencies. 

Despite its promising results, several avenues for future research remain. First, one could explore more sophisticated and automated methods for constructing the neighborhood subsets. While effective, the current pre-defined partitioning scheme based on K-Means could be replaced by a dynamic or even learnable partitioning strategy, which might allow the model to adapt its local attention scope more flexibly. Second, a natural extension is to evaluate the model's performance on longer-term prediction tasks, such as forecasting several hours ahead. This would provide a more challenging benchmark to assess the model's capacity for capturing exceptionally long-range dependencies. Finally, the core principles of the STGAtt framework are generalizable. Applying the model to other complex spatial-temporal domains, such as weather prediction, crowd flow dynamics, or financial time series analysis, represents a promising direction for future work.

In conclusion, STGAtt presents a significant advancement in traffic flow forecasting by effectively modeling intricate spatial-temporal relationships, offering a promising avenue for future research in this domain.

% \bibliographystyle{plain}
% \bibliography{main}

\vfill

\end{sloppypar}
\end{document}